\documentclass{IEEEtran}
\usepackage{array}
\usepackage{textcomp}
\usepackage{stfloats}
\usepackage{url}
\usepackage{verbatim}
\usepackage{cite}
\usepackage{booktabs}
\usepackage{hyperref}
\usepackage{subfigure}
\usepackage{times}
\usepackage{soul}
\usepackage{xcolor}         % colors
\usepackage{graphicx}
\usepackage{amsmath}
\usepackage{amsthm}
\usepackage{amssymb,amsfonts}
\usepackage{algpseudocode}
\usepackage[linesnumbered,ruled,vlined]{algorithm2e}
\usepackage{changepage}
\usepackage{multirow}

\newtheorem{theorem}{Theorem}
\newtheorem{lemma}{Lemma}
\newtheorem{assumption}{Assumption}
\newtheorem{definition}{Definition}
\DeclareUnicodeCharacter{FF0C}{,}
\hypersetup{
    colorlinks=false,       
    linkcolor=blue,       
    citecolor=blue,       
    filecolor=magenta, 
    urlcolor=cyan         
}
\begin{document}

\title{Enhancing DP-SGD through Non-monotonous Adaptive Scaling Gradient Weight}

\author{Tao Huang,~\IEEEmembership{}
		Qingyu Huang,~\IEEEmembership{}
		Xin Shi,~\IEEEmembership{}
            Jiayang Meng,~\IEEEmembership{}
            Guolong Zheng,~\IEEEmembership{}
		Xu Yang,~\IEEEmembership{}
            Xun Yi，~\IEEEmembership{}

\thanks{Tao Huang, Qingyu Huang, Xin Shi, Xu Yang, and Guolong Zheng are with the School of Computer Science and Big Data, Minjiang University, Fuzhou City, Fujian Province, China, e-mail: huang-tao@mju.edu.cn, qingyuhuang@stu.mju.edu.cn, shixin@stu.mju.edu.cn, yangxu9111@gmail.com, gzheng@mju.edu.cn.}
\thanks{Jiayang Meng is with the School of Information, Renmin University of China, Haidian District, Beijing, e-mail: jiayangmeng@ruc.edu.cn.}
\thanks{Xun Yi is with the RMIT University, Melbourne, Australia, e-mail: xun.yi@rmit.edu.au.}
}
% The paper headers
\markboth{IEEE TRANSACTIONS ON DEPENDABLE AND SECURE COMPUTING, VOL. 21, NO. 4, JULY/AUGUST 2024}%
{Shell \MakeLowercase{\textit{et al.}}: A Sample Article Using IEEEtran.cls for IEEE Journals}

% \IEEEpubid{0000--0000/00\$00.00~\copyright~2021 IEEE}
% Remember, if you use this you must call \IEEEpubidadjcol in the second
% column for its text to clear the IEEEpubid mark.

\maketitle
\begin{abstract}
In the domain of deep learning, the challenge of protecting sensitive data while maintaining model utility is significant. Traditional Differential Privacy (DP) techniques such as Differentially Private Stochastic Gradient Descent (DP-SGD) typically employ strategies like direct or per-sample adaptive gradient clipping. These methods, however, compromise model accuracy due to their critical influence on gradient handling, particularly neglecting the significant contribution of small gradients during later training stages.
In this paper, we introduce an enhanced version of DP-SGD, named Differentially Private Per-sample Adaptive Scaling Clipping (DP-PSASC). This approach replaces traditional clipping with non-monotonous adaptive gradient scaling, which alleviates the need for intensive threshold setting and rectifies the disproportionate weighting of smaller gradients. Our contribution is twofold. First, we develop a novel gradient scaling technique that effectively assigns proper weights to gradients, particularly small ones, thus improving learning under differential privacy. Second, we integrate a momentum-based method into DP-PSASC to reduce bias from stochastic sampling, enhancing convergence rates. Our theoretical and empirical analyses confirm that DP-PSASC preserves privacy and delivers superior performance across diverse datasets, setting new standards for privacy-sensitive applications.
\end{abstract}


\section{Introduction}
\IEEEPARstart{D}{eep} networks have achieved great access recently. However, the training of these networks relies heavily on large quantities of high-quality data, which often includes sensitive personal information, making data-driven deep models vulnerable to privacy attacks, such as membership inference attacks \cite{b1,b2}, reconstruction attacks \cite{b3,b4,b5}, property inference attacks \cite{b6,b7} and so on. In the realm of deep learning, ensuring privacy while maintaining the utility of deep neural models has emerged as a pivotal challenge. 

\begin{figure}[htbp]
\centering
\includegraphics[scale = 0.06]{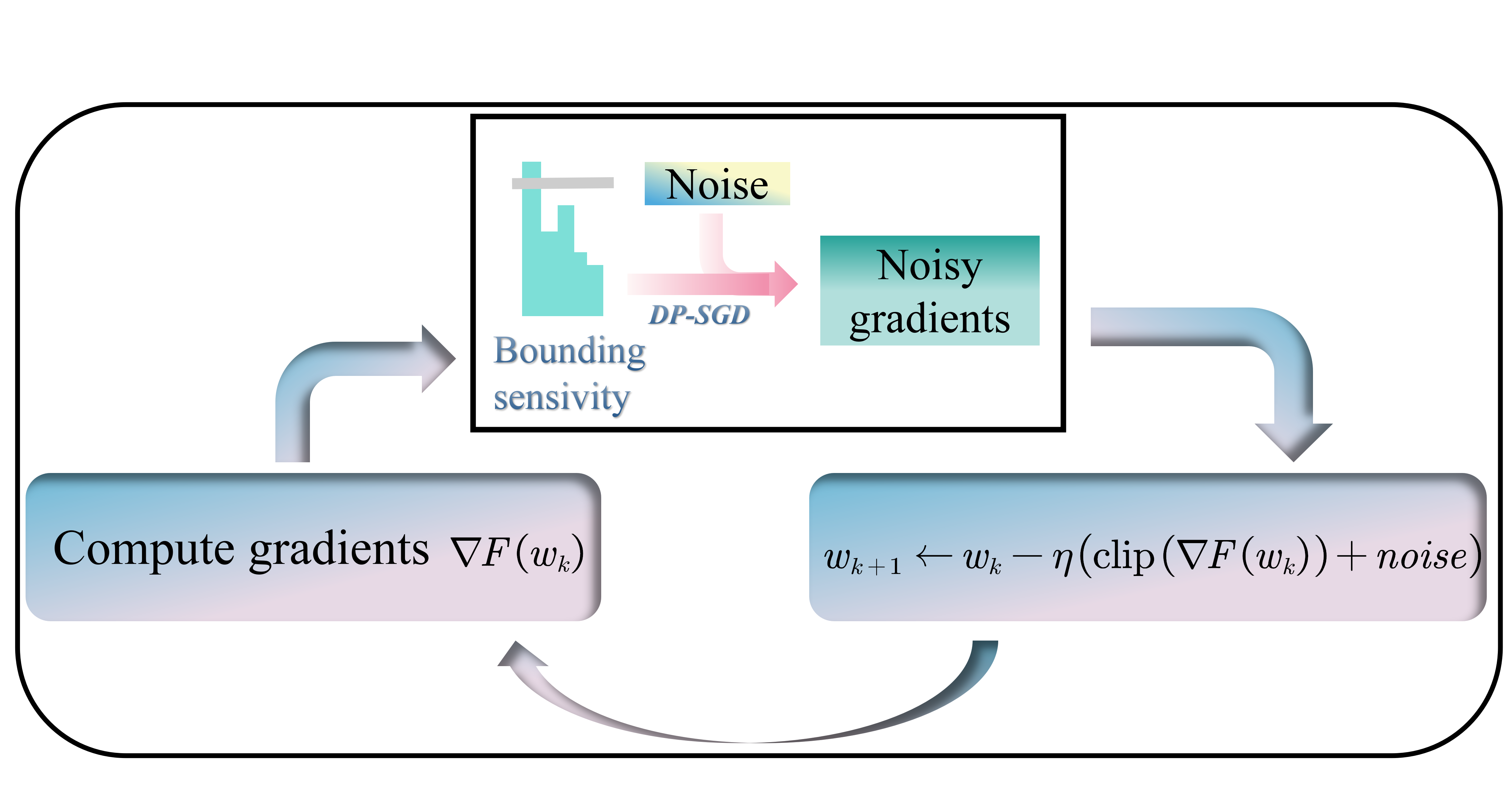}
\caption{Differentially Private Stochastic Gradient Descent (DP-SGD)}
\label{02}
\end{figure}

Differential Privacy (DP) \cite{b8}, a framework for safeguarding sensitive data, is widely adopted in deep learning. Stochastic gradient descent (SGD) combined with DP, usually denoted DP-SGD \cite{b9}, has been demonstrated effective in providing rigorous guarantees, as illustrated in Fig. \ref{02}. Different from SGD, the iteration of DP-SGD when updating $w_k$ to $w_{k+1}$ is $w_{k+1} \leftarrow w_k - \frac{\eta}{B} \left( \sum_{i \in \mathcal{B}_k} {g}_{k,i} \cdot \min \{ 1, \frac{C}{\|{g}_{k,i}\|} \} + \sigma \cdot C \cdot \mathcal{N} \left( 0, I \right) \right)$, where $w$ is the model parameter, $\eta$ is the learning rate, $\mathcal{B}_k$ is the random batch in $k$-th iteration, $C$ is a predefined constant to constrain the sensitivity of the gradients $g_{k,i}$ and $\sigma$ is the standard deviation of differentially private Gaussian noise. Namely, the gradients $g_{k,i}$ are first clipped by the bound $C$ and then are injected noise before updating $w$.

Though effective, DP-SGD relies heavily on gradient clipping to meet privacy guarantees. The accuracy of deep models, however, is significantly influenced by the choice of the clipping threshold $C$, a parameter that determines the maximum allowable influence of each data sample on the model's updates. This threshold plays a central role in balancing the trade-off between privacy and accuracy. Specifically, an improperly set clipping threshold can lead to either excessive noise addition, which obscures useful data patterns and degrades model accuracy, or insufficient privacy protection, which risks exposing sensitive data features. As such, the choice of clipping threshold directly impacts the utility of models trained under differential privacy constraints, with profound implications for their deployment in sensitive applications. Some workable methods are to consume an extra privacy budget to search the proper $C$ \cite{b10} or estimate the optimal $C$ \cite{b11,b12}. However, these methods either incur a heavier privacy load or lead to new privacy leaking problems.

One promising approach to circumvent the labor-intensive process of setting a clipping threshold is to replace the clipping operation with strategic gradient scaling. Strategic gradient scaling not only avoids searching or setting a clipping threshold but also eliminates the bias introduced by directly clipping the gradient \cite{b16}. The original strategic gradient scaling is automatic clipping/normalizing \cite{b13,b14}. Automatic clipping/normalizing constrains the privacy sensitivity by scaling the original gradient $g$ to $\tilde{g} = \frac{g}{\| g \| +r}$ where $r$ is a small constant to make DP-SGD stable. All per-sample gradients are normalized to the same magnitude since $\| \tilde{g} \| \leq 1$. Consequently, the batch gradient becomes a weighted average of the per-sample gradients and the scaling weight is $\frac{1}{\| g \| +r}$. Automatic clipping/normalizing suffers from that smaller gradients are given larger weights. In the extreme case, the scaling weight of small gradients is $\frac{1}{r}$ which is a very large number. To improve automatic clipping/normalizing, DP-PSAC \cite{b15} proposes the modified scaling weight $C / \left(\|g\|+\frac{r}{\|g\|+r} \right)$. The modified scaling weight aims to give small gradients with small weights to avoid their excessive influences on the weight update while not influencing privacy sensitivity. However, DP-PSAC exhibits significant limitations, particularly in how it handles small gradients. DP-PSAC often assigns disproportionately small weights to small gradients, particularly in the later stages of training. This approach can inadvertently minimize the contribution of such gradients, stifling the model's ability to refine its parameters and converge to an optimal solution. Moreover, DP-PSAC ignores the bias introduced by stochastic sampling in SGD. The bias has non-negligible impacts on the convergence rate of DP-SGD \cite{b17}. Noise addition can only mitigate the privacy risks but can not eliminate the bias \cite{b18,b19}. 

To overcome these limitations, we enhance DP-PSAC from two aspects. First, we modify the scaling weight to assign proper weights to small gradients since we find that small gradients are important when updating the model's parameters in the later training stages. Second, we aim to eliminate the bias introduced by stochastic sampling via the momentum method \cite{b20} to get an improved convergence rate. Based on these, we propose DP-SGD integrated with a new non-monotonous gradient scaling weight, called \textbf{D}ifferentially \textbf{P}rivate \textbf{P}ersample \textbf{A}daptive \textbf{S}caling \textbf{C}lipping (shorten by DP-PSASC). Moreover, by combining with the momentum method, we propose the momentum version of DP-PSASC with an improved convergence rate of models. We summarize our contributions as follows:

\begin{itemize}

    \item We introduce a new non-monotonous gradient scaling weight that assigns proper weights to small gradients, especially in the later training stages. Based on this, we propose DP-PSASC, an improved DP-SGD method as private as the currently used DP-SGD.

    \item We combine the momentum method with DP-PSASC and eliminate the bias introduced by the stochastic sampling. The momentum version of DP-PSASC has an improved convergence rate compared to DP-PSAC.

    \item We demonstrate the theoretical and empirical superiority of the proposed algorithms. Through extensive experiments, we demonstrate the new state-of-the-art performance of DP-PSASC learning over several datasets.
\end{itemize}

\section{Related work}

\textbf{Deep learning and differential privacy.} Differentially private learning with gradient clipping and the Gaussian mechanism has emerged as a leading method in deep learning. The concept of constant clipping was first introduced by \cite{b9} to integrate privacy protection into SGD, a method known as DP-SGD. This approach was further explored in subsequent studies \cite{b21,b22,b23,b24,b25,b26,b27,b28,b29} to apply DP to other optimization algorithms, including DP-AdaGrad, DP-SVRG, and ApolySFW. From a theoretical standpoint, Zhang et al. \cite{b30} and Zhang et al. \cite{b31} examined the convergence properties of clipped SGD. On the application side, DP-Lora \cite{b32} and RGP \cite{b33} facilitated differential privacy in large-scale model fine-tuning using techniques like low-rank compression.

\textbf{Dynamic threshold adjustments.} Research indicates that the optimal clipping threshold varies throughout the optimization process \cite{b34}. Several studies have proposed dynamically adjusting the threshold during training to mitigate the drawbacks of a fixed threshold on DP-based algorithm performance. Andrew et al. \cite{b11} estimated the optimal clipping threshold using an additional privacy budget during optimization. Du et al. \cite{b35} suggested progressively decreasing the clipping threshold and noise magnitude over iterative rounds. More granular methods, such as those by Pichapati et al. \cite{b36} and Asi et al. \cite{b37}, introduced axis-specific adaptive clipping and noise addition, applying different clipping thresholds and non-uniform noise to gradient components on different axes. Despite these advancements, these methods still require manual setting of the initial threshold, and the final performance is highly sensitive to this initial value.

\textbf{Gradient scaling based approaches.} To address the dependency on a clipping threshold, Bu et al. \cite{b13} and Yang et al. \cite{b14} concurrently proposed using normalization to control gradient sensitivity, termed Automatic Clipping (Auto-S) or Normalized SGD (NSGD). These approaches demonstrated that by normalizing all gradients to a uniform magnitude, only one hyperparameter needs tuning as the learning rate and clipping parameter become coupled. However, this technique introduces significant deviation between the normalized batch-averaged gradient and the original when some gradient norms are very small. To improve automatic clipping/normalizing, DP-PSAC \cite{b15} proposes the non-monotonous adaptive weight for gradients. The modified scaling weight aims to give small gradients with small weights while not influencing privacy sensitivity. However, this approach can inadvertently minimize the contribution of such gradients, stifling the model's ability to refine its parameters and converge to an optimal solution. Moreover, DP-PSAC ignores the bias introduced by stochastic sampling in SGD. Our method and algorithms overcome these limitations by assigning proper weights to small gradients in the later training stages and introducing the momentum method to improve the convergence rate. The proposed approaches achieve better theoretical and experimental results. 

\section{Preliminary}

\subsection{Notations and definitions}
The important notations are in TABLE \ref{notations}. Consider a deep neural network $F(w)$ where $w$ is the network parameter to be optimized. Given a private dataset $\mathcal{D}=\{ d_1,\ldots, d_N \}$ where $d_i = (x_i,y_i)$, the loss function $f(d_i,w)$ is the empirical loss between the output of the deep neural network and the ground truth. The training process is to find $w$ satisfying $w = \arg \min_{w} F(w) = \arg \min_{w} \frac{1}{N} \sum_{i=1}^N f(d_i,w)$. The optimization problem is usually denoted Empirical Risk Minimization (ERM). Below, we formally introduce the definitions of Lipschitz continuity and smoothness, which are commonly used in
optimization research.

\begin{definition}[\textbf{Lipschitz continuity}] The function $F(w)$ is $L$-Lipschitz continuous if for all $w_1, w_2$ in the domain, $| F(w_1) - F(w_2)| \leq L \| w_1 - w_2 \|$.
\end{definition}

\begin{definition}[\textbf{Smoothness}]
    The function $F(w)$ is $\beta$-smooth if $\nabla F(w)$ is $\beta$-Lipschitz continuous. 
\end{definition}

\begin{definition}[\textbf{Differential privacy} \cite{b8}]
A randomized mechanism $\mathcal{M}$ satisfies $(\epsilon, \delta)$-differential privacy (DP) if for any two adjacent datasets $D$ and $D^{\prime}$ that differ by a single individual's data, and for all $S \subseteq \text{Range}(\mathcal{M}) $(the set of all possible outputs of $\mathcal{M}$, the following inequality holds:
    $\Pr[\mathcal{M}(D) \in S] \leq e^\epsilon \cdot \Pr[\mathcal{M}(D') \in S] + \delta$
\end{definition}

\begin{lemma}[\textbf{Gaussian Mechanism for Differential Privacy} \cite{b8}]
Let $ \mathcal{M}: \mathcal{D} \rightarrow \mathbb{R}^k $ be a function with  $\ell_2$-sensitivity $\Delta_2 \mathcal{M} = \| \mathcal{M}(D) - \mathcal{M}(D^{\prime}) \|$ which measures the maximum change in the Euclidean norm of $\mathcal{M}$ for any two adjacent datasets $D$ and $D^{\prime}$ that differ by a single individual's data. The Gaussian Mechanism adds noise drawn from $\mathcal{N}(0, \sigma^2 I)$ to the output of $\mathcal{M}$ and provides $(\epsilon, \delta)$-differential privacy if $\sigma \geq \frac{\Delta_2 \mathcal{M} \cdot \sqrt{2 \log(1.25 / \delta)}}{\epsilon}$.
\label{gaussian}
\end{lemma}

\subsection{DP-SGD}
Differential privacy is often combined with SGD and widely applied in deep learning to safeguard both model and data privacy. The method is called DP-SGD \cite{b9}. The core idea of DP-SGD involves perturbing gradients by injecting handcrafted Gaussian noise. Given the original gradients $\{{g}_i\}_{i \in \mathcal{B}}$ where $\mathcal{B}$ represents a batch set of private samples, DP-SGD updates $w$ using the perturbed gradients as follows:
\begin{equation}
w \leftarrow w - \frac{\eta}{B} \left ( \sum_{i \in \mathcal{B}} {g}_i \cdot \min \{ 1, \frac{C}{\|{g}_i\|} \} + \sigma \cdot C \cdot \mathcal{N} \left( 0, I \right) \right )
\label{abadi_eq}
\end{equation} 
Since $\| {g}_i \cdot \min \{ 1, \frac{C}{\|{g}_i\|} \}  \| \leq C$, the sensitivity of the clipped gradients is $C$. According to Lemma \ref{gaussian}, Eq.(\ref{abadi_eq}) satisfies $(\epsilon, \delta)$-DP if we choose $\sigma \geq \frac{\sqrt{2 \log(1.25 / \delta)}}{\epsilon}$. DP-SGD often encounters issues due to the improper selection of $C$. To enhance DP-SGD, one could choose an appropriate $C_t$ for each iteration $t$ or $C_L$ for each layer in the model. Alternatively, a grid search for the learning rate $\eta$ and $C$ could be conducted. However, grid searches can consume extra privacy budgets and be computationally intensive.

\begin{table}
\caption{Symbols and Notations}
\begin{tabular}{l|l}
\toprule
\textbf{Symbols} & \textbf{Notations} \\ \midrule
$\| \cdot \|$       & $\ell_2$ norm of a vector or spectral norm of a matrix         \\ \hline
$\langle \cdot , \cdot \rangle$        &  the inner product of two vector         \\ \hline
$\mathbb{E} [\cdot]$        &     a random variable's mathematical expectation      \\ \hline
$\Pr[\cdot]$        &   probability that some event occurs        \\ \hline
\multirow{2}{*}{$\mathcal{D} = \{ d_1, \ldots, d_N \}$}        &   the private training dataset with $N$ training     \\ 
 & samples\\\hline
\multirow{2}{*}{$F(w)$}        &   the empirical risk loss concerning the model's     \\ 
& parameter $w$\\\hline
\multirow{2}{*}{$f(w,d_i)$}    & the empirical loss function concerning the private   \\ 
                            & data sample $d_i$\\ \hline
$\nabla f$ or $ \nabla F$        &  the gradient of $f$ or $F$         \\ \hline
$w_k$        & the model's parameter in $k$-th epoch          \\ \hline
$g (g_{k,i})$        &  the original gradient of $d_i$ in $k$-th epoch       \\ \hline
$\tilde{g} (\tilde{g}_{k,i})$        & the scaled gradient of $d_i$ in $k$-th epoch          \\ \hline
$\hat{g} (\hat{g}_{k})$        &  the noisy gradients in $k$-th epoch         \\ \bottomrule
\end{tabular}
\label{notations}
\end{table}
\subsection{Gradient scaling weight}
\label{sub2.3}

In Eq.(\ref{abadi_eq}), the coefficient $\min \{ 1, \frac{C}{||{g}_i||} \}$ can be regarded as the gradient scaling weight $w({g}_i)$. The designer can set appropriate gradient scaling weights to bound the gradient's sensitivity and avoid setting or searching the proper $C$. Usually, the weight $w({g}_i)$ is designed as a function of the original gradients. Thus, the principle of designing $w({g}_i)$ is to analyze the properties of the gradients to improve the performance of DP-SGD.

The weight function, $w({g}_i) = \min \{ 1, \frac{C}{\|{g}_i\|} \}$ \cite{b9}, has two significant drawbacks. When a large proportion of gradients satisfies $\|{g}_i\| \geqslant C$, many gradients will be clipped. Conversely, when $\|{g}_i\| \leqslant C$ and particularly when the norm is very small, the noise generated from the Gaussian distribution $\mathcal{N} \left( 0, I \right)$ can be much greater than the original gradients. In both cases, the performance of the DP-SGD can be compromised.

To address these issues, automatic clipping/normalizing (Auto-S/NSGD) is proposed, introducing a gradient scaling weight \cite{b13,b14}:
\begin{equation}
w({g}_i) = \frac{1}{\|{g}_i\|+r}
\end{equation}
, where $r$ is a small regularization term to enhance training stability. Consequently, DP-SGD updates $w$ using the perturbed gradients as follows:
\begin{equation}
w \leftarrow w - \frac{\eta}{B} \left ( \sum_{i\in \mathcal{B}} \frac{ {g}_i}{\|{g}_i\| + r} + \sigma \cdot \mathcal{N} \left( 0, I \right) \right )
\label{auto_eq}
\end{equation}
In Eq.(\ref{auto_eq}), it is easy to verify that $\| \frac{ {g}_i}{\|{g}_i\| + r} \| \leq 1$. Therefore, the gradient sensitivity is also $1$. Automatic clipping/normalizing makes it possible to adaptively scale the influence of each sample based on its norm. 
However, the gradient weight in Eq.(\ref{auto_eq}) assigns larger weights to smaller gradients, potentially increasing the weighted gain of a sample by up to $\frac{1}{r}$ times when its gradient approaches zero, where $r$ is typically set to 0.01 or less. Small gradient samples, however, have minimal impact on batch-averaged gradients for the entire batch. Therefore, the monotonic weight function in Eq.(\ref{auto_eq}) can sometimes be inappropriate, leading to lower accuracy of the trained models.

To further improve the performance, a non-monotonic weight function is proposed in Differentially Private Per-Sample Adaptive Clipping (DP-PSAC) \cite{b15}:

\begin{equation}
w({g}_i) = \frac{C}{\|{g}_i\|+\frac{r}{\|{g}_i\|+r}}
\label{DP-PSAC}
\end{equation}

This non-monotonic adaptive weight avoids assigning very large weights to small gradient samples, thereby reducing convergence errors. It assigns weights near 1 to small gradient samples, while weighting large gradients similarly to $\frac{1}{\|{g}_i\|}$. DP-SGD updates $w$ using the perturbed gradients as follows:

\begin{equation}
w \leftarrow w - \frac{\eta}{B} \left ( \sum_{i\in \mathcal{B}} \frac{C \cdot {g}_i}{\|{g}_i\|+\frac{r}{\|{g}_i\|+r}} + \sigma \cdot C \cdot \mathcal{N} \left( 0, I \right) \right )
\label{pasc_eq}
\end{equation}

From Eq.(\ref{pasc_eq}), we can see that $C$ can be integrated into the learning rate of the original optimizer, eliminating the need for grid search operations. Consequently, no additional privacy budget is required for tuning $C$.

\section{Our Method}
In developing differentially private stochastic gradient descent (DP-SGD), balancing robust privacy with sustained model performance presents a formidable challenge. DP-PSAC introduces an innovative non-monotonic adaptive weight that dynamically adjusts the gradient weights based on gradient norms to enhance privacy-preserving mechanisms in deep learning. While this approach significantly improves upon traditional DP-SGD, it also harbors intrinsic limitations that can adversely affect model performance and the integrity of privacy guarantees. Previous work has overlooked crucial aspects of managing gradient magnitudes during training, particularly in the later stages where gradients significantly influence the effectiveness of privacy guarantees.

\subsection{Ignored aspects in DP-PSAC}  

\begin{figure}
    \centering
    \vspace{-2mm}
    \subfigure[Gradient Norm Distribution(FashionMNIST)]{
		\includegraphics[width=0.4\textwidth]{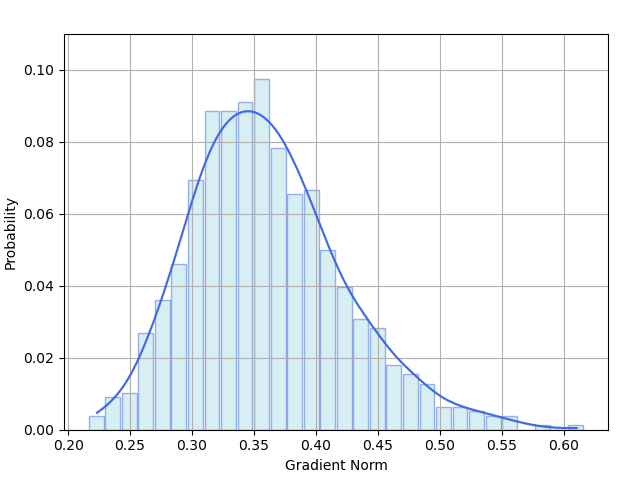}
		\label{grad_norm_pro1}
    }\\
    \subfigure[Gradient Norm Distribution(CelebA)]{
		\includegraphics[width=0.4\textwidth]{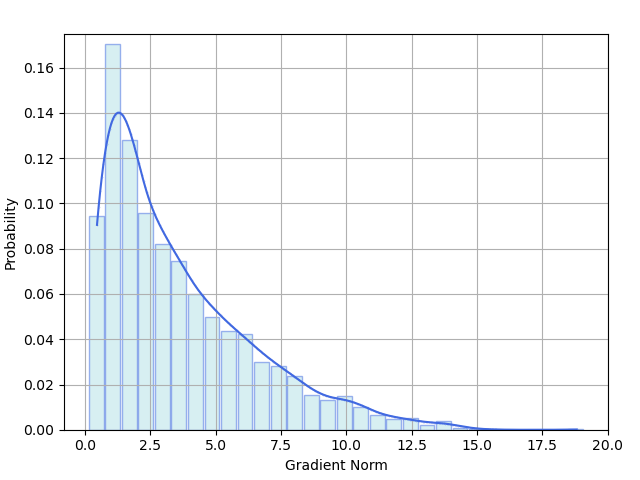}
		\label{grad_norm_pro2}
    }
    \caption{Gradient Norm Distribution In The Last 10 Epochs}
    \label{grad}
\end{figure}

\textbf{High proportion of small gradients:} In the later training stages, there is an increased proportion of small gradients across batches. This shift suggests a transition to a fine-tuning phase and signals that the model is approaching a local minimum, where subtle yet critical adjustments are essential for optimal performance. We analyze and present the empirical distribution of gradients relative to their norms in the last 10 epochs over a CNN network trained with the FashionMNIST dataset and a ResNet9 network trained with the CelebA dataset. Fig. \ref{grad_norm_pro1} and Fig. \ref{grad_norm_pro2} display a shift distribution towards small gradients, indicating their increased impact on parameter updates during later training stages. This underscores the need to give greater consideration to these smaller gradients, which attracted less attention in the previous works.

By neglecting these factors, DP-PSAC tends to undervalue smaller gradients by assigning them disproportionately lower weights. This approach can skew the learning process, potentially ignoring minor but significant features (represented by smaller gradients), and thus deviating from the intended learning trajectory. Such bias can diminish the model's generalizability from training data to unseen data, compromising both accuracy and robustness.

\textbf{Impact of noise in DP-SGD.} Traditional differential privacy optimizers incorporate noise addition to obscure the influence of individual data samples, protecting against privacy breaches via gradient analysis. For example, in DP-SGD, the original gradient $g$ is masked by adding differentially private noise $n$, expressed as $\hat{g} = g + n$. To enhance privacy, one goal of DP-SGD is to maximize the ratio $\frac{\| n \|}{\| g \|}$ under the same privacy level, thereby more effectively masking the private information contained in $g$. DP-PSAC offers potential for further improvement in this regard that can achieve greater privacy without necessitating additional privacy budget expenditure.

Addressing these critical challenges necessitates the introduction of a new gradient scaling function within DP-SGD. By adjusting the scale of gradients according to their magnitudes, we hope that the proposed function would ensure that even as gradients diminish, they remain capable of contributing effectively to model learning, thereby enhancing both the practical utility and theoretical robustness of privacy-preserving machine learning models.

\subsection{Non-monotonous Scaling Gradient Weight}

\textbf{Pay more attention to small gradients. \label{more}} To enhance performance through the utilization of a non-monotonous weight function, we introduce a refined model termed "Non-Monotonous Scaling Weighted Gradients" based on the empirical pieces of evidence, as delineated in Eq.(\ref{scaling}). This model employs a revised weighting scheme to optimize outcomes effectively:

\begin{equation}
    \tilde{g}=\mathrm{Clip}\left( g \right) =\frac{C\cdot g}{s\cdot \|g\|+\frac{r}{\|g\|+r}}.
    \label{scaling}
\end{equation}

In the revised formulation, the weight of the original gradient, denoted as $w_{s}(g)$, is defined as $w_{s}(g) = C / \left(s\cdot\|g\|+\frac{r}{\|g\|+r} \right)$, where $s$ represents the scaling coefficient of gradients. 

By introducing $s$ and comparing Eq.(\ref{scaling}) with Eq.(\ref{DP-PSAC}), the new method provides an additional lever to control how much influence the norm of the gradient has on the clipping process. This can be particularly beneficial in situations where the distribution of gradient norms is highly skewed or varies significantly during training. To understand this, we can see that adjusting $s$ allows for finer control over the relative weighting of smaller gradients, potentially reducing the bias towards smaller gradients. For detailed explorations, the maximum value of $w_{s}(g)$ is given by $\max w_s\left( g \right)  =C / \left( 1-\left( 1-\sqrt{sr} \right) ^2 \right)$. As $s$ increases, $\max w_s\left( g \right) $ decreases. This implies that, with a larger $s$, the weights of small gradients in the later stages of training can be effectively limited and minimized. When $s=1$, Eq.(\ref{scaling}) degrades to Eq.(\ref{DP-PSAC}). Thus, we can find a proper $s<1$ for different datasets to assign proper and larger weights to small gradients. The analysis also reveals the existence of an optimal scaling factor, $s^*$, which signifies the peak performance of the differentially private model. Moreover, $\lim_{\| g \| \rightarrow 0} \frac{w_s\left( g \right)}{C} = 1$. That is the same as DP-PSAC and it means the proposed weight is not too small or large and will not produce improperly scaled gradients for model training.

By scaling gradients inversely with their magnitude through the term $\frac{C \cdot g}{s \cdot \|g \| + \frac{r}{\|g\| + r}}$, we enhance the relative influence of smaller gradients by not diminishing their magnitude excessively. This scaling not only preserves the direction and integrity of the gradient but also balances the signal-to-noise ratio, improving the stability and convergence of the model under privacy constraints. A scaling function that modulates based on gradient magnitude addresses this distribution shift effectively. It ensures that small but numerous gradients are not nullified by aggressive normalization or swamped by noise. By amplifying the contribution of these small gradients relatively, the proposed function facilitates a more nuanced update process, preserving delicate training dynamics that are essential for achieving high accuracy in tasks requiring fine discrimination. Consequently, this approach helps maintain an optimal balance between privacy and learning efficacy, allowing the model to continue learning meaningful patterns from the data without being excessively perturbed by the privacy-preserving mechanisms.

\textbf{Amplify the impact of noise under the same level of privacy guarantee. \label{amplify}} Consider comparing the addition of noise to a gradient using our method and DP-PSAC. For an original gradient $g$, the noise-augmented gradient in DP-PSAC is $g^* = \frac{C\cdot g}{\|g\|+\frac{r}{\|g\|+r}} + n^*$ where $n^* \sim \mathcal{N}(0, \sigma^2 C^2 I)$. Under the same level of privacy guarantee, to fulfill the criteria for $(\epsilon, \delta)$-differential privacy (DP), the norm of the scaled gradient obtained via our method satisfies: $\| \frac{C\cdot  g }{s\cdot \|g\|+\frac{r}{\|g\|+r}} \| \leq \frac{C\cdot\|g\|}{s \cdot \|g\|}=\frac{C}{s}$. Consequently, the scale of noise can be appropriately adjusted to $\frac{C \cdot \sigma}{s}$, with $\sigma$ being the same as in DP-PSAC. The noise-augmented gradient then becomes: $g^{**} = \frac{C\cdot g}{s \cdot \|g\|+\frac{r}{\|g\|+r}} + n^{**}$ where $n^{**} \sim \mathcal{N}(0, \frac{\sigma^2 C^2}{s^2} I)$. Then, we have:

\begin{equation}
    \begin{split}
        \frac{\mathbb{E} \| n^{**} \|}{\| g^{**} \|} \geq \frac{\mathbb{E} \| n^{*} \|}{\| g^{*} \|} & \Leftrightarrow \frac{\frac{C \sigma \sqrt{d}}{s}}{\| \frac{C\cdot  g }{s\cdot \|g\|+\frac{r}{\|g\|+r}} \|} \geq \frac{C \sigma \sqrt{d}}{\| \frac{C\cdot g}{\|g\|+\frac{r}{\|g\|+r}} \|}\\
        & \Leftrightarrow \frac{\| g \|}{\|g\|+\frac{r}{\|g\|+r}} \geq \frac{s \cdot \| g \|}{s \cdot \|g\|+\frac{r}{\|g\|+r}}\\
        &\Leftrightarrow \|g\|+\frac{r}{\|g\|+r} \leq \|g\|+\frac{\frac{r}{s}}{\|g\|+r} ,
    \end{split}
\end{equation}
which holds when $s \leq 1$. This indicates that our method can more effectively obscure private gradients under the same privacy guarantee level.

\subsection{Algorithms and Theoretical Analysis}

Based on the above analysis, we propose DP-SGD integrated with non-monotonous scaling weight gradients, called \textbf{D}ifferentially \textbf{P}rivate \textbf{P}ersample \textbf{A}daptive \textbf{S}caling \textbf{C}lipping (shorten by DP-PSASC), in Algorithm \ref{alg: algorithm1}.

\begin{algorithm*}[htb]
    \caption{DP-PSASC}
    \label{alg: algorithm1}
    \KwIn{Initial Weights $w_0$, learning rate $\eta$, Batch Size $B$, Dataset $\mathcal{D}=\{ d_1,\ldots, d_N \}$, Privacy Budget $(\epsilon, \delta)$, Maximum Norm Size $C$, Number of Iterations $T$, Scaling Coefficient $s$.} 
     Compute the noise scale $\sigma$ \\
    \For{$k=0;k<T;k++$}{
         Sample a batch $\mathcal{B}_{k} = \{ d_i \}_{i=1}^B$ from $\mathcal{D}$ uniformly without replacement \\
         Compute the gradient $g_{k,i}$ for each sample $d_i \in \mathcal{B}_{k}$ \\
         $\tilde{g}_{k,i}=\frac{C \cdot g_{k,i}}{s\cdot \|g_{k,i}\|+\frac{r}{\|g_{k,i}\|+r}}$ \\
         $\hat{g}_k=\sum_{i=1}^B{\tilde{g}_{k,i}}+\mathcal{N}\left( 0,\frac{C^2\sigma ^2}{s^2} \right)$ \\
         $w_{k+1} \leftarrow w_{k} - \frac{\eta}{B} \hat{g}_k$ \\
        }
\end{algorithm*}

From a theoretical perspective, adjusting $s$ provides a way to achieve the same privacy guarantees with less impact on model performance while maintaining or even improving convergence rates. The flexibility offered by $s$ allows for more nuanced adjustments to how gradients are clipped, which can lead to better optimization of the privacy-accuracy trade-off. For example, by tuning $s$ based on the specific characteristics of the dataset and the learning dynamics, one can achieve higher model accuracy without substantially compromising privacy. In this part, we divide the privacy guarantee and the convergence analysis of our proposed method.

The first thing we want to emphasize is that $C$ is avoided to be searched and could be integrated into the learning rate. It is because the model increment in the $k$-th iteration can be formulated as:
\begin{equation}
    \begin{aligned}
   \Delta w_k&=-\frac{\eta}{B}\left( \sum_{i=1}^B{\tilde{g}_{k,i}}+N\left( 0,\frac{C^2\sigma ^2}{s^2} \right) \right)\\
   &=-\frac{\eta \cdot C}{B}\left( \sum_{i=1}^B{\frac{g_{k,i}}{s\cdot ||g_{k,i}||+\frac{r}{||g_{k,i}||+r}}}+N\left( 0,\frac{\sigma ^2}{s^2} \right) \right)
   \end{aligned}
\end{equation}

For the privacy guarantee of DP-PSASC, we have the following theorem which can be derived from any common privacy analysis for DP-SGD. 

\begin{theorem}[\textbf{Privacy Guarantee of DP-PSASC}] There exist constants $c_1$ and $c_2$ so that given the number of iterations $T$, for any $\epsilon \leq c_1 T (\frac{B}{N})^2$ and $\delta \leq \frac{1}{N}$, Algorithm \ref{alg: algorithm1} is $(\epsilon, \delta)$-differentially private if $\sigma \geq c_2 \frac{B \sqrt{T \ln (1/\delta)}}{N \epsilon}$.
\end{theorem}

The next issue is whether DP-PSASC converges. We point out that DP-PSAC suffers from a bias in its convergence rate (see Theorem 2 in \cite{b15}) for they ignore the bias introduced by stochastic sampling in SDG (presented in our assumption \ref{assump1}). Generic theoretical analyses of DP-SGD with a scaling operation are presented for smooth non-convex optimization. We make minimal assumptions on the stochastic gradient distribution and only assume its second moment is bounded. To make the analysis more practicable, we also make an assumption concerning the norm of the per-sample gradient when the training process moves on. These assumptions are formally stated below.

\begin{assumption}[\textbf{Bounded Second Moment of Stochastic Gradient}] For any sample $(x, y)$ drawn from the dataset $\mathcal{D}$, the expected squared norm of the deviation between the full gradient and the stochastic gradient is bounded by $\tau ^2$, formally:
\begin{equation}
    \mathbb{E}_{(x,y) \sim \mathcal{D}}\left[ \|\nabla F \left( w \right) - \nabla f \left( w, x, y \right)\|^2 \right] \leqslant \tau ^2.
\end{equation}
\label{assump1}
\end{assumption}

\begin{assumption}[\textbf{Decreasing Gradient Norm Per Sample}] Assume that for each sample $(x_i,y_i) \in \mathcal{D}$, the norm of the gradient $\| g_{k,i} \| = \| \nabla f (w_k,x_i,y_i) \|$ monotonically decreases through the iterations, i.e.,
\begin{equation}
    \| g_{k,i} \| \leq \| g_{k-1,i} \|,
\end{equation}
implying improved optimization behavior as training progresses.
\label{assump2}
\end{assumption}

\begin{theorem}[\textbf{Convergence Guarantee of DP-PSASC}] Under the assumptions, if $F$ is $\beta$-smooth, then there exist some constant $\hat{k}$ and for $r = O(T^{-\frac{1}{2}})$, $\eta=\sqrt{\frac{2 s^2 F\left( w_{\hat{k}} \right)}{\beta (T-\hat{k}) (2+\sigma^2 d)}}$, DP-PSASC ensures that for $k=1,2,...,T$, $\mathbb{E} _k\left[ \min _k||\nabla F\left( w_{k-1} \right) || \right] \leqslant O(T^{-\frac{1}{2}} + \tau)$.
\label{convergence1}
\end{theorem}

The main source of bias in DP-SGD comes from the stochastic nature of the sampled gradients that are used to update the model's parameters. Namely, SGD introduces the bias $\tau$ which is presented in Theorem \ref{convergence1} .To mitigate the impacts caused by the sampling bias, the momentum-based variance reduction method is often adopted \cite{b20}. The momentum method, by averaging gradients over past iterations, can smooth out the effects of noisy or biased gradients. When gradients are clipped in DP-SGD, the stochastic nature of sampling can introduce bias, particularly if the clipped gradients are noisy or unevenly distributed. Momentum helps by averaging these gradients, reducing their variance and the overall noise, which indirectly corrects for biases introduced by stochastic sampling. 

DP-PSASC can be integrated with the momentum-based method smoothly. The integrated methods are presented in Algorithm \ref{alg: algorithm2} and Algorithm \ref{alg: algorithm3} where Algorithm \ref{alg: algorithm2} is the gradient descent (GD) version and Algorithm \ref{alg: algorithm3} is the stochastic gradient descent (SGD) version. For Algorithm \ref{alg: algorithm3}, we adopt a specific technique called inner-outer momentum \cite{b17}. In this approach, per-sample scaled gradients over past iterates are first exponentially averaged (inner momentum) before scaling. After scaling, these gradients are then averaged again over past iterates (outer momentum) before being applied in the differentially private gradient descent update. This two-fold averaging process further smooths out erratic fluctuations in the gradient values due to stochastic sampling, leading to a more stable and accurate optimization path.

We demonstrate that the sampling bias of DP-PSASC is successfully eliminated and the convergence guarantee is presented in Theorem \ref{convergence2}.  

\begin{algorithm*}[htb]
    \caption{DP-PSASC with momentum (GD version)}
    \label{alg: algorithm2}
    \KwIn{Initial Weights $w_0$, learning rate $\eta$, Dataset $\mathcal{D}=\{ d_1,\ldots, d_N \}$, Privacy Budget $(\epsilon, \delta)$, Maximum Norm Size $C$, Momentum Parameter $\gamma$, Number of Iterations $T$, Scaling Coefficient $s$.} 
     Compute the noise scale $\sigma$ \\
     Initialize $m_{0,i} = g_{0,i}$\\
    \For{$k=1;k<T;k++$}{
         Compute the gradient $\{g_{k,i}\}_{i=1}^{N}$ \\
         $\tilde{g}_{k,i}=\frac{C \cdot m_{k,i}}{s\cdot \|m_{k,i}\|+\frac{r}{\|m_{k,i}\|+r}}$ \\
         Compute $g_k = \sum_{i=1}^{N} \tilde{g}_{k,i}$\\
         $\hat{g}_k=g_k+\mathcal{N}\left( 0,\frac{C^2\sigma ^2}{s^2} \right)$ \\
         $w_{k+1} \leftarrow w_{k} - \frac{\eta}{B} \hat{g}_k$\\
         \For{$i=1;i<N;i++$}{$m_{k,i} = \gamma g_{k,i} + (1-\gamma)m_{k-1,i}$\\}
        }
\end{algorithm*}

\begin{algorithm*}[htb]
    \caption{DP-PSASC with momentum (SGD version)}
    \label{alg: algorithm3}
    \KwIn{Initial Weights $w_0$, learning rate $\eta$, Batch Size $B$, Dataset $\mathcal{D}=\{ d_1,\ldots, d_N \}$, Privacy Budget $(\epsilon, \delta)$, Maximum Norm Size $C$, Inner Momentum Parameter $\gamma_0$, Outer Momentum Parameter $\gamma_1$, Momentum length $K_0$, Number of Iterations $T$, Scaling Coefficient $s$.} 
     Compute the noise scale $\sigma$ \\
     Initialize outer momentum $M_0 = 0$\\
    \For{$k=1;k<T;k++$}{
        Sample a batch $\mathcal{B}_{k} = \{ d_i \}_{i=1}^B$ from $\mathcal{D}$ uniformly without replacement \\
        Compute the gradient $\{g_{k,i}\}_{i=1}^{B}$ \\
        For each $d_i \in \mathcal{B}_{k}$, compute inner per-sample momentum $m_{k-1,i} = \sum_{l=k-1-K_0}^{k-1} \gamma_{0}^{k-1-l} g_{l,i}$\\
         $\tilde{g}_{k,i}=\frac{C \cdot m_{k,i}}{s\cdot \|m_{k,i}\|+\frac{r}{\|m_{k,i}\|+r}}$ \\
         Compute $g_k = \sum_{i=1}^{B} \tilde{g}_{k,i}$\\
         Compute outer momentum $M_k = (1-\gamma_1)M_{k-1} + g_k + \mathcal{N}\left( 0,\frac{C^2\sigma ^2}{s^2} \right)$\\
         $w_{k+1} \leftarrow w_{k} - \frac{\eta}{B} M_k$\\
        }
\end{algorithm*}

\begin{theorem}[\textbf{Convergence Guarantee of DP-PSASC with Momentum}] Under the conditions of Theorem \ref{convergence1}, for $r = O(T^{-\frac{1}{2}})$, $s = O(T^{-\frac{1}{4}})$, $\gamma = O(T^{-\frac{1}{4}})$, $\eta=O(T^{-\frac{3}{4}})$, DP-PSASC with momentum ensures that for $k=1,2,...,T$, $\mathbb{E} _k\left[ \min _k\|\nabla F\left( w_{k-1} \right) \| \right] \leqslant O(T^{-\frac{1}{4}})$.
\label{convergence2}
\end{theorem}

From Theorem \ref{convergence2}, we can see that by setting proper scaling coefficient $s$ and momentum parameter $\gamma$, the bias introduced by stochastic sampling can be eliminated. Compared with Theorem 2 in DP-PSAC, we obtain a vanishing upper bound $O(T^{-\frac{1}{4}})$ and the term independent of $O(T^{-\frac{1}{4}})$ is eliminated.

\section{Experiments}
\subsection{Experimental Settings}

\textbf{Model settings.} We evaluate performances over four datasets and models. We train a CNN model consisting of four convolutional layers and one fully connected layer with MNIST \cite{b41} and FashionMNIST \cite{b42}. For the CIFAR10 dataset \cite{b43}, we train a simCLR model we keep the same experimental setup as \cite{b38} and use pre-trained SimCLRv2 \cite{b39} based on contrastive learning.  Further, we train a ResNet9 model \cite{b40} on Imagenette \cite{b44} to validate the performance of our method on more complex multi-classification and multilabel classification problems. While conducting experiments, we keep $C$ and $r$ in Eq.(\ref{abadi_eq}), Eq.(\ref{auto_eq}), Eq.(\ref{pasc_eq}) and Eq.(\ref{scaling}) the same. We find the proper $s$ in Eq.(\ref{scaling}) to implement our proposed DP-PSASC to get higher test accuracy. After finding the proper $s$, we fix $s$ and implement the momentum version of DP-PSASC with the proper momentum parameters. The momentum version of DP-PSASC is simply denoted as DP-PSASC*. The hyperparameter setting can be referred to in TABLE \ref{hyper}.

\begin{table*}
\centering
\small
\caption{Hyperparameter Settings}
\begin{tabular}{c|cccp{2cm}p{2cm}c}
\toprule
\multirow{3}{*}{\textbf{Datasets and Models}} & \multicolumn{5}{c}{\textbf{Hyperparameter Values}} \\ \cline{2-7} 
                          & \multirow{2}{*}{$\| \mathcal{D} \|$}   & \multirow{2}{*}{Batch-size $B$}   & \multirow{2}{*}{Epoch $T$}   & Stable   & Sensivity  & \multirow{2}{*}{$(\epsilon,\delta)$}
                          \\ 
                          & && & Parameter $r$ &Bound $C$ &\\\midrule
\textit{MNIST(CNN)}                     & 60,000    & 512         & 20     & 0.0001 & 0.3 & $(3,1e-5)$    \\ \hline
\textit{FashionMNIST(CNN)}             & 40,000        & 512         & 60     & 0.001   & 0.25 &$(9,1e-5)$   \\ \hline
\textit{CIFAR10(SimCLR)}                   & 50,000        & 1024             & 40         & 0.0001     & 0.1 & $(2,1e-5)$      \\ \hline
\textit{CelebA([Male])(ResNet9)}                   & 162,079        & 512             & 10         & 0.0001    & 0.2 & $(8,5e-6)$     \\ \hline
\textit{Imagenette(ResNet9)}                & 9,464        & 256             & 50         & 0.01    & 0.2  & $(8,1e-4)$    \\ \bottomrule
\end{tabular}
\label{hyper}
\end{table*}

\textbf{Baselines and configurations.} We compare our method with DP-SGD (Eq.(\ref{abadi_eq})) \cite{b9}, Auto-S/NSGD (Eq.(\ref{auto_eq})) \cite{b13,b14}, and DP-PSAC (Eq.(\ref{pasc_eq})) \cite{b15}. All experiments are performed on a server with two Intel(R) Xeon(R) Silver 4310 CPUs @2.10GHz, and four NVIDIA 4090 GPUs. The operating system is Ubuntu 22.04 and the CUDA Toolkit version is 12.4. All computer vision experimental training procedures are implemented based on the latest versions of Pytorch and Opacus \cite{b45}.

\textbf{Evaluation metrics.} 
This study concentrates on the model accuracy within the context of differential privacy, particularly investigating the implications of two primary aspects: gradient variation under differential privacy and gradient scaling weight in the later training stages. All experiments are conducted 5 times and the test accuracy is the average result.

% \textit{Parameter Similarity:} Initially, we establish a baseline by training the model without the implementation of differential privacy, denoting its parameters as $w_{\text{np}}$. Subsequent models are trained using various configurations of differentially private stochastic gradient descent (DP-SGD). The parameters from these models, denoted $w$, are reshaped into vectors to facilitate comparison. We quantify the similarity between the differentially private parameters and the non-private parameters $w_{\text{np}}$ by calculating their cosine similarity. This metric helps assess how DP-SGD methods offer privacy guarantees for the model's parameters. 

\textit{Gradient similarity:} We then examine the influence of injected noise on the accuracy of gradient computations during the training process. For this analysis, we focus on the gradients computed in the last 10 epochs. Let $g_{i,j}$ represent the gradient corresponding to the $j$-th batch in the $i$-th epoch with $g_{i,j}^{\text{p}}$ and $g_{i,j}^{\text{np}}$ indicating the noisy (private) and original (non-private) gradients, respectively. The cosine similarity between $g_{i,j}^{\text{p}}$ and the batch-average gradient $\frac{1}{|\mathcal{B}_{j}|} \sum_{j \in \mathcal{B}_{j}} g_{i,j}^{\text{np}}$ is calculated to evaluate the deviation introduced by privacy-preserving mechanisms. The range of similarities obtained is divided into uniform intervals to determine the distribution probability of gradients within these intervals, thereby illustrating the variability of gradient estimation introduced by DP-SGD across different training batches.

\textit{Gradient scaling weight:} We analyze the average gradient scaling weight derived from DP-PSAC and our newly proposed DP-PSASC method. Specifically, we calculate the gradient scaling weights for all gradients in the final 10 epochs and document their average. This analysis aims to examine the variation in gradient scaling weights during the later stages of training and to explore the relationship between these variations and the model's performance. This relationship underscores the significant impact that small gradient norms have on the scaling weight, thereby influencing the model's training dynamics and overall performance. We exclude results pertaining to Auto-S/NSGD (Eq.(\ref{auto_eq})) \cite{b13,b14} from this analysis. In the later phases of training, the gradient scaling weights of Auto-S/NSGD become excessively large, which can be attributed to the gradients' diminished norms. As discussed in subsection \ref{sub2.3}, when the norm of the gradients is small, the scaling weight tends to approach $\frac{1}{r}$, where $r$ is a small value.

We hope these analyses provide insights into the trade-offs between model accuracy and privacy preservation in DP-SGD, highlighting the extent to which privacy mechanisms influence parameter estimation and gradient fidelity.

\subsection{Experiment Results}
\textbf{Model's accuracy.} 
As shown in TABLE \ref{acc}, our proposed DP-PSASC achieves higher test accuracy compared with DP-SGD, Auto-S/NSGD, and DP-PSAC. For different datasets, the scaling weights are a bit different. We contribute the improved performance to the appropriate scaling weights that we have implemented in the later stages of training. Actually, $s$ has little impact in the initial training stage but it becomes significantly influential in the later stages. Furthermore, DP-PSASC*, the momentum-enhanced version of DP-PSASC, demonstrates superior performance compared to its non-momentum counterpart. Specifically, the inclusion of momentum effectively mitigates the bias introduced by stochastic sampling, as we demonstrated in Theorem \ref{convergence2}.

\begin{table*}
\centering
\small
\caption{Test accuracy of DP-SGD, Auto-S, DP-PSAC, and our methods}
\begin{tabular}{c|ccccc}
\toprule
\multirow{2}{*}{\textbf{Datasets and Models}} & \multicolumn{5}{c}{\textbf{Differentially Private Methods}} \\ \cline{2-6} 
                          & DP-SGD   & Auto-S/NSGD   & DP-PSAC   & DP-PSASC  & DP-PSASC* 
                          \\ \midrule
\textit{MNIST(CNN)}                     & 97.35    & 97.95         & 98.11     & 98.37($s=0.9$)  & 98.65($s=0.9$)      \\ \hline
\textit{FashionMNIST(CNN)}             & 85.59        & 86.15         & 86.36     & 86.91($s=0.55$)   & 87.22($s=0.55$)    \\ \hline
\textit{CIFAR10(SimCLR)}                   & 92.24        & 92.65             & 92.85         & 93.12($s=0.8$)     & 93.36($s=0.8$)       \\ \hline
\textit{CelebA([Male])(ResNet9)}                   & 94.79        & 95.12             & 95.2         & 95.36($s=0.45$)     & 95.61(s=0.45)       \\ \hline
\textit{Imagenette(ResNet9)}                & 63.61        & 63.87             & 64.21         & 65.26($s=0.8$)    & 66.79($s=0.8$)       \\ \bottomrule
\end{tabular}
\label{acc}
\end{table*}

\begin{figure*}[htp!]
    \centering
    \vspace{-2mm}
    \subfigure[DP-SGD v.s Ours]{
		\includegraphics[width=0.3\textwidth]{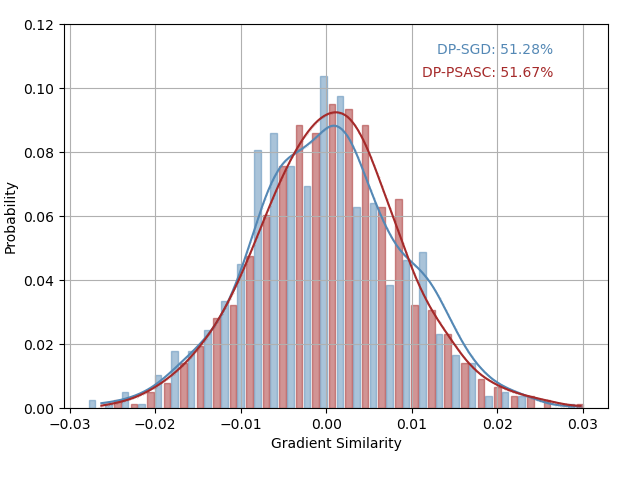}
		\label{grad_DPSGD_f}
    }
    \subfigure[Auto-S/NSGD v.s Ours]{
		\includegraphics[width=0.3\textwidth]{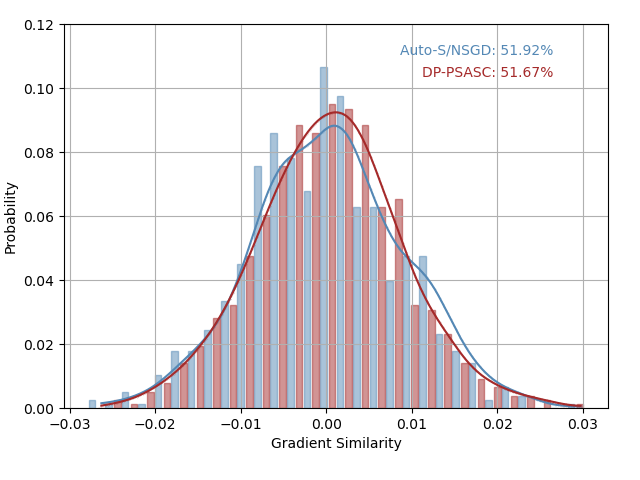}
		\label{grad_Auto_f}
    }
    \subfigure[DP-PSAC v.s Ours]{
		\includegraphics[width=0.3\textwidth]{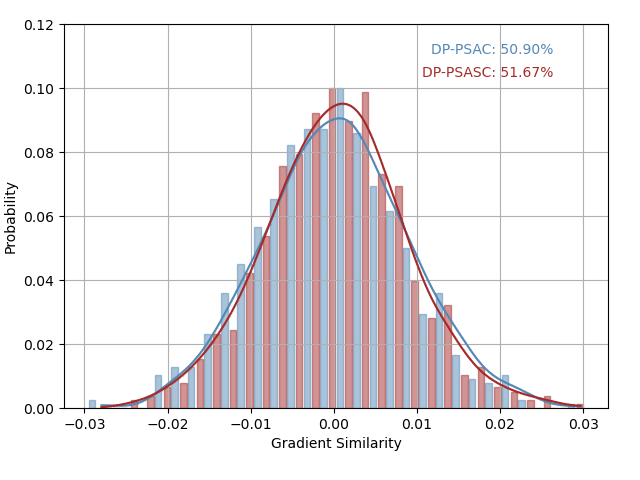}
		\label{grad_DPPSAC_f}
    }
    \caption{Gradient Similarities(FashionMNIST)}
    \label{fmnist}
\end{figure*}

\begin{figure*}[htp!]
    \centering
    \vspace{-2mm}
    \subfigure[DP-SGD v.s Ours]{
		\includegraphics[width=0.3\textwidth]{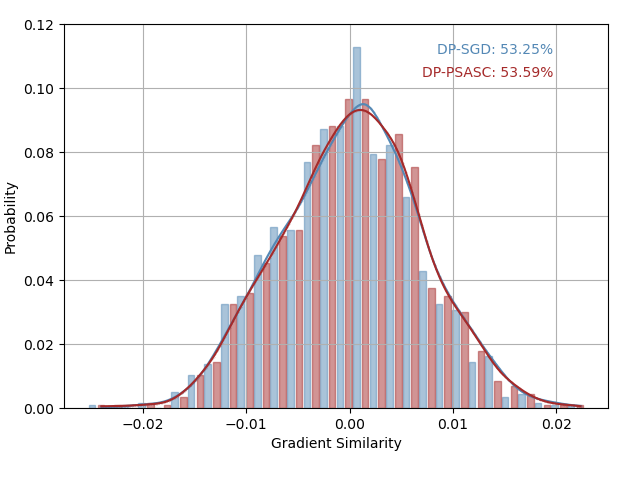}
		\label{grad_DPSGD_m}
    }
    \subfigure[Auto-S/NSGD v.s Ours]{
		\includegraphics[width=0.3\textwidth]{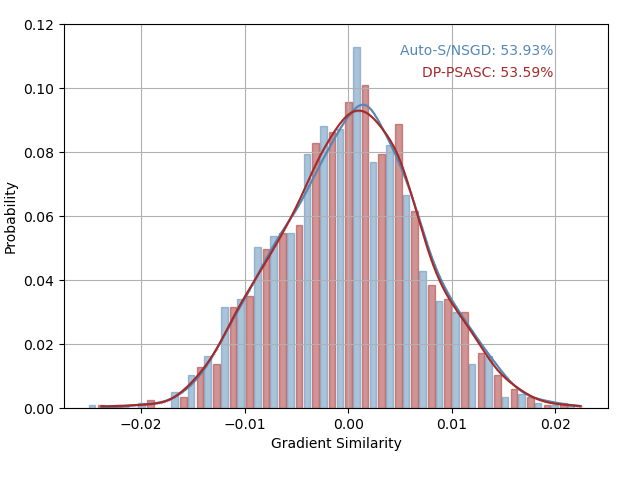}
		\label{grad_Auto_m}
    }
    \subfigure[DP-PSAC v.s Ours]{
		\includegraphics[width=0.3\textwidth]{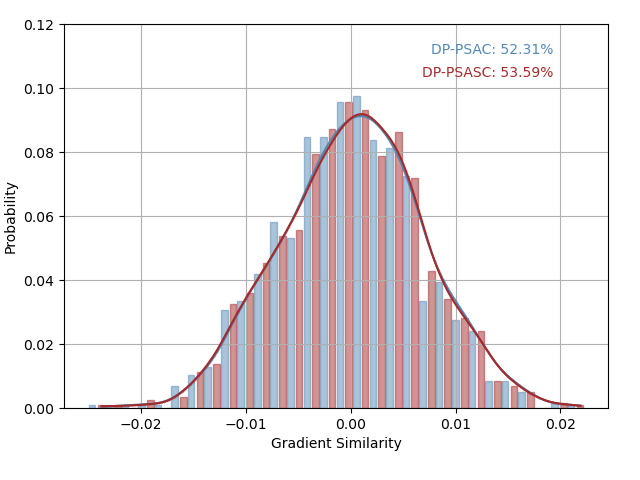}
		\label{grad_DPPSAC_m}
    }
    \caption{Gradient Similarities(MNIST)}
    \label{mnist1}
\end{figure*}

\begin{figure*}[htp!]
    \centering
    \vspace{-2mm}
    \subfigure[DP-SGD v.s Ours]{
		\includegraphics[width=0.3\textwidth]{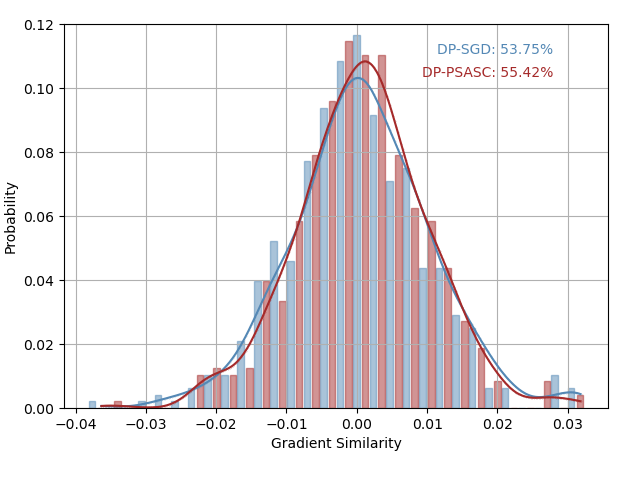}
		\label{grad_DPSGD_ci}
    }
    \subfigure[Auto-S/NSGD v.s Ours]{
		\includegraphics[width=0.3\textwidth]{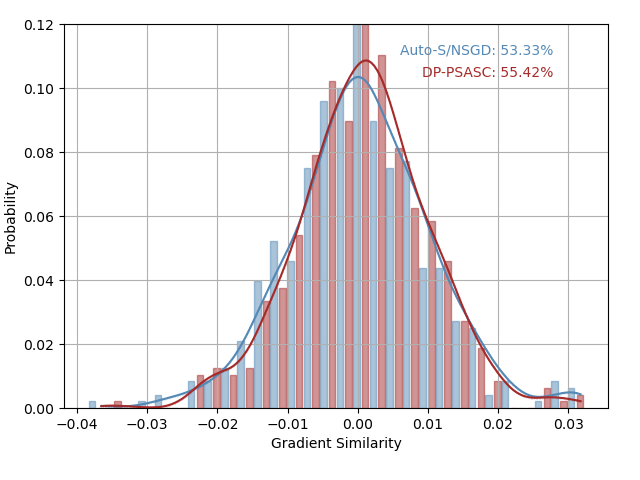}
		\label{grad_Auto_ci}
    }
    \subfigure[DP-PSAC v.s Ours]{
		\includegraphics[width=0.3\textwidth]{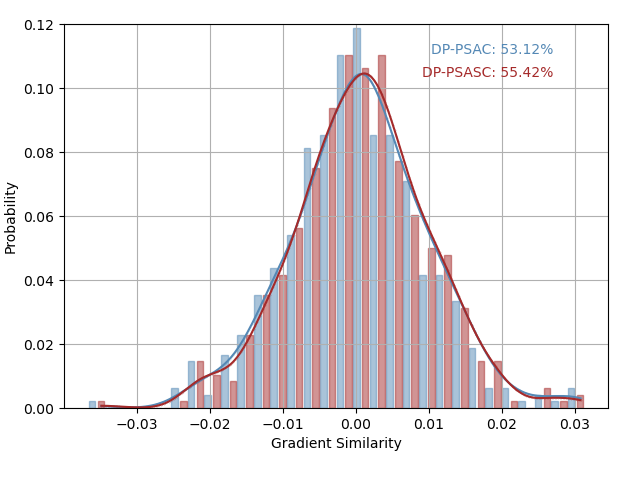}
		\label{grad_DPPSAC_ci}
    }
    \caption{Gradient Similarities(CIFAR10)}
    \label{mnist2}
\end{figure*}

\begin{figure*}[htp!]
    \centering
    \vspace{-2mm}
    \subfigure[DP-SGD v.s Ours]{
		\includegraphics[width=0.3\textwidth]{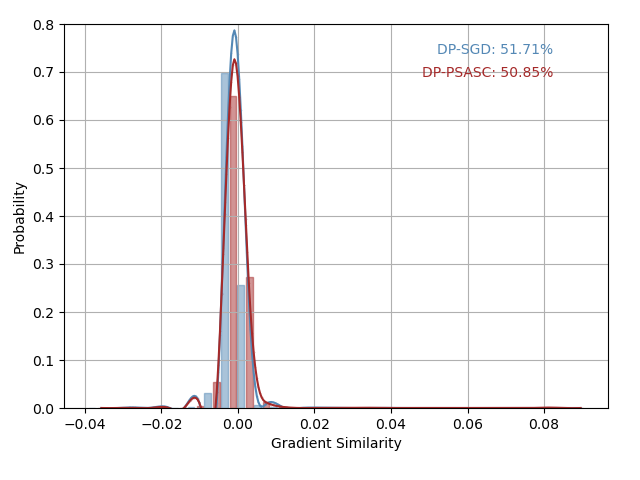}
		\label{grad_DPSGD_ce}
    }
    \subfigure[Auto-S/NSGD v.s Ours]{
		\includegraphics[width=0.3\textwidth]{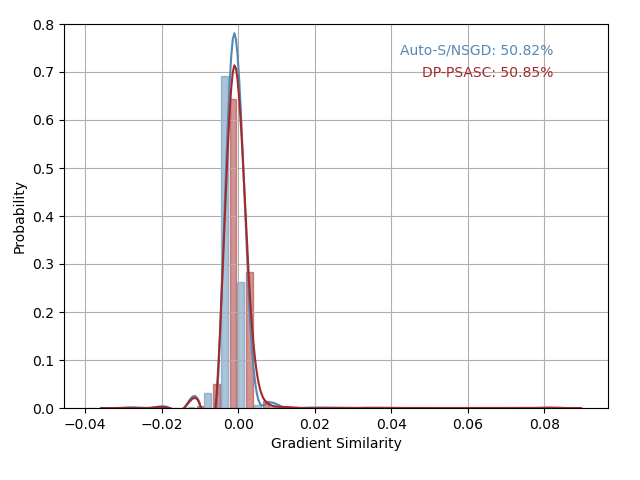}
		\label{grad_Auto_ce}
    }
    \subfigure[DP-PSAC v.s Ours]{
		\includegraphics[width=0.3\textwidth]{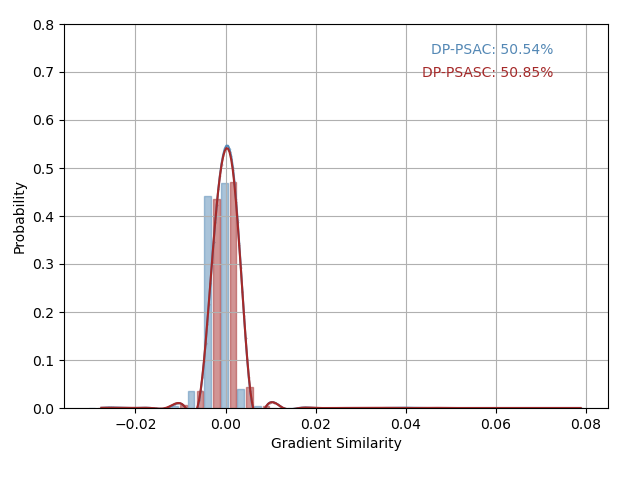}
		\label{grad_DPPSAC_ce}
    }
    \caption{Gradient Similarities(CelebA[Male])}
    \label{mnist3}
\end{figure*}

\begin{figure*}[htp!]
    \centering
    \vspace{-2mm}
    \subfigure[DP-SGD v.s Ours]{
		\includegraphics[width=0.3\textwidth]{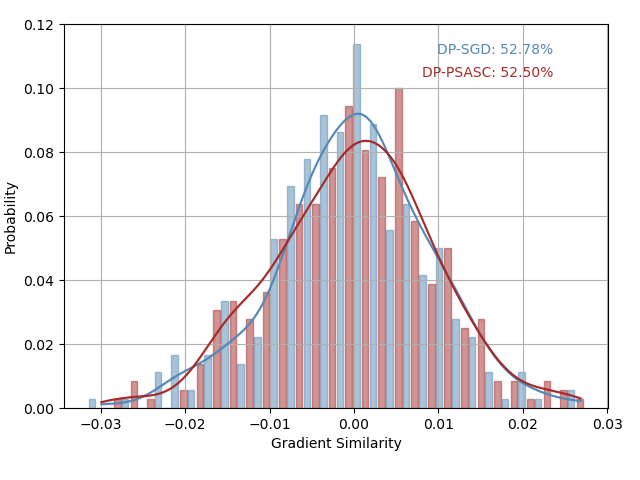}
		\label{grad_DPSGD_i}
    }
    \subfigure[Auto-S/NSGD v.s Ours]{
		\includegraphics[width=0.3\textwidth]{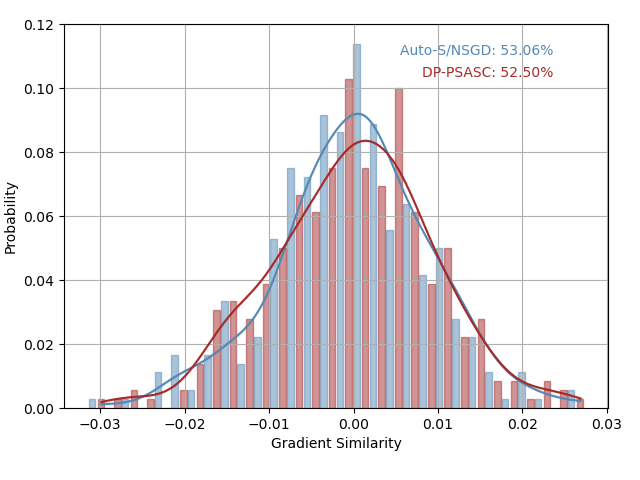}
		\label{grad_Auto_i}
    }
    \subfigure[DP-PSAC v.s Ours]{
		\includegraphics[width=0.3\textwidth]{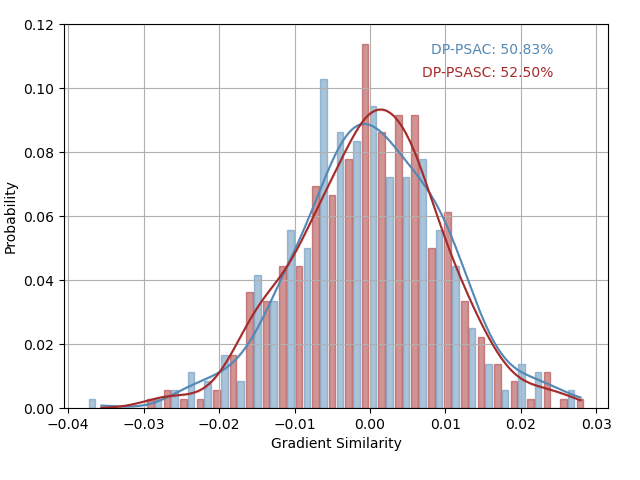}
		\label{grad_DPPSAC_i}
    }
    \caption{Parameter Similarities and Gradient Similarities(Imagenette)}
    \label{Imagenette}
\end{figure*}

\begin{figure*}[htp!]
    \centering
    \vspace{-2mm}
    \subfigure[FashionMNIST]{
		\includegraphics[width=0.29\textwidth]{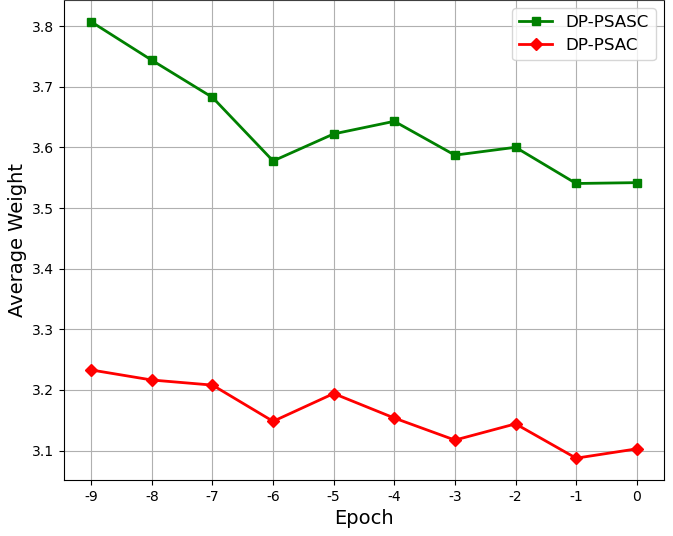}
		\label{scaling_f}
    }
    \subfigure[MNIST]{
		\includegraphics[width=0.29\textwidth]{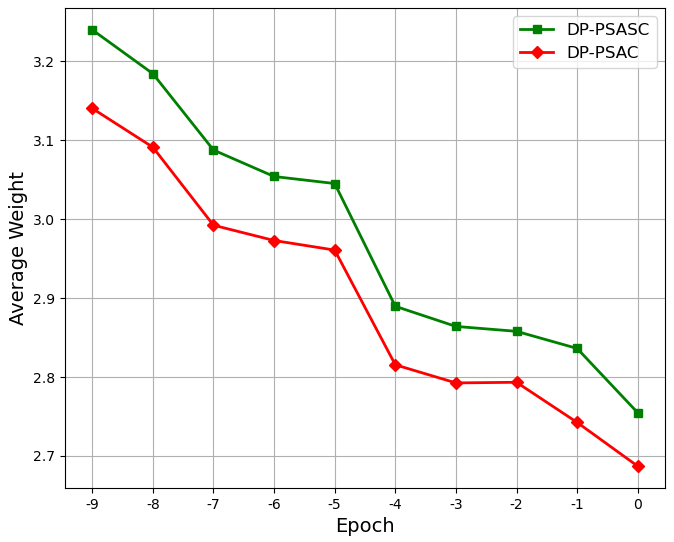}
		\label{scaling_m}
    }
    \subfigure[CIFAR10]{
		\includegraphics[width=0.29\textwidth]{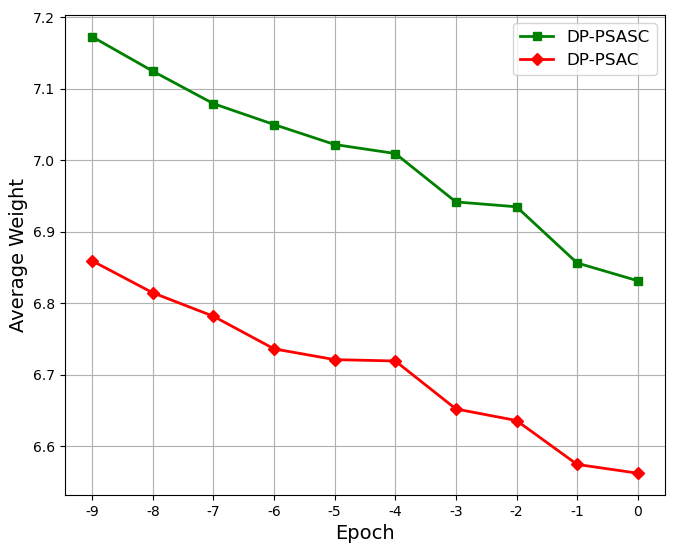}
		\label{scaling_ci}
    }\\
    \subfigure[CelebA]{
		\includegraphics[width=0.28\textwidth]{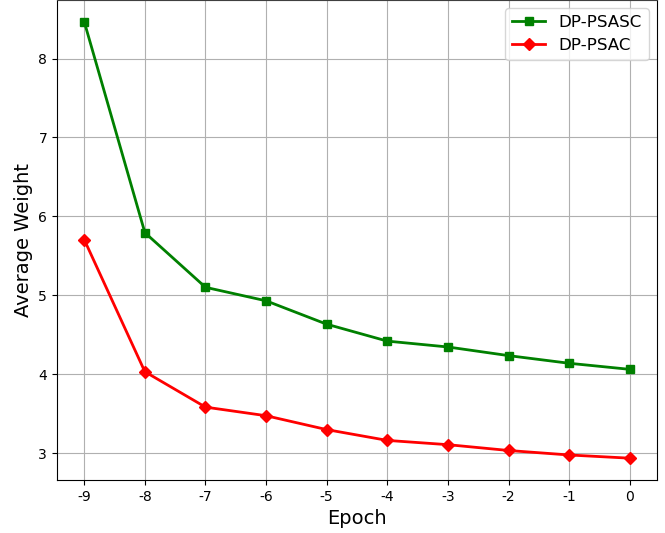}
		\label{scaling_ce}
    }
    \subfigure[Imagenette]{
		\includegraphics[width=0.29\textwidth]{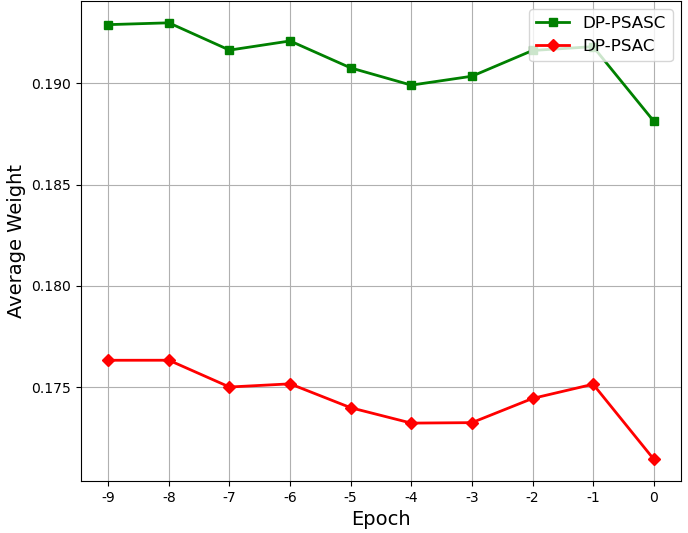}
		\label{scaling_imag}
    }
    \caption{Gradient Scaling Weights}
    \label{scaling_fig}
\end{figure*}

% When calculating the cosine similarity, we first transform parameters into flat vectors. The smaller the cosine similarity indicates the angle of two vectors is bigger, namely these two vectors are less similar. Figure \ref{cos_para_f} illustrates the cosine similarity between parameters derived by various differentially private methods and non-private parameters. The parameters derived through our method exhibit smaller similarity to the non-private parameters, outperforming those obtained via traditional DP-SGD,  Auto-S/NSGD, and DP-PSAC. This demonstrates our analysis about the amplification of the impact of noise under the same level of privacy guarantee in subsection \ref{amplify}. Moreover, when the training process iterates over epochs, the cosine similarity of all methods becomes smaller. This indicates that differentially private noise accumulates as the training process goes on. Thus, we need to set a proper number of epochs to avoid the collapse of the obtained model. 

\textbf{Gradient similarity.} To calculate the cosine similarity, gradients are first transformed into flattened vectors. A lower cosine similarity value indicates a larger angle between two vectors, implying less similarity between them. In the final 10 epochs of training, the proportion of gradients in our proposed DP-PSASC with larger cosine similarities exceeds that of gradients in DP-PSAC, as illustrated in Fig. \ref{grad_DPPSAC_f}, Fig. \ref{grad_DPPSAC_m}, Fig. \ref{grad_DPPSAC_ci}, Fig. \ref{grad_DPPSAC_ce}, and Fig. \ref{grad_DPPSAC_i}. This suggests that gradients from DP-PSASC more closely resemble the non-private gradients compared to those from DP-PSAC, particularly in the later stages of training. This is in line with our focus on smaller gradients in the later training stages, discussed in subsection \ref{more}.

\begin{figure*}[htp]
    \begin{adjustwidth}{-0.25in}{-0.25in}  % 设置1/4英寸的左右边距
        \centering
        \includegraphics[scale=0.15 ]{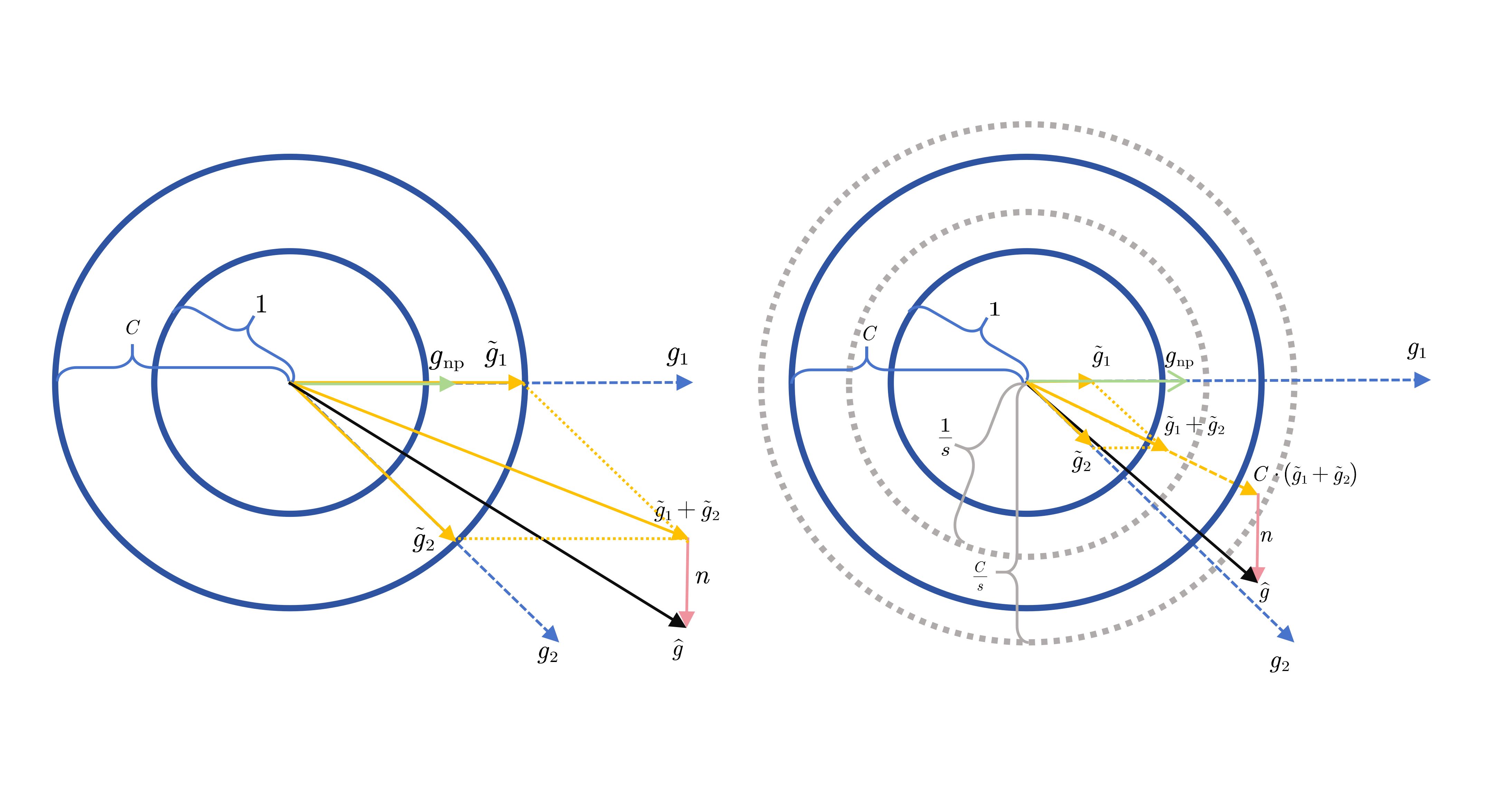}  
        \caption{Explainations on Gradient Similarity}
        \label{explaination}
    \end{adjustwidth}
\end{figure*}

An interesting observation emerges from the results shown in Fig. \ref{grad_DPSGD_f} and \ref{grad_Auto_f}, Fig. \ref{grad_DPSGD_m} and \ref{grad_Auto_m}, Fig. \ref{grad_DPSGD_ci} and \ref{grad_Auto_ci}, Fig. \ref{grad_DPSGD_ce} and \ref{grad_Auto_ce}, Fig. \ref{grad_DPSGD_i} and \ref{grad_Auto_i}. In the last 10 epochs, the proportion of gradients in DP-PSASC that exhibit larger cosine similarities is sometimes either equal to or smaller than that in traditional DP-SGD and Auto-S/NSGD. For an explanation of this phenomenon, refer to Fig. \ref{explaination}. Cosine similarity measures the angle between two vectors. In traditional DP-SGD and Auto-S/NSGD, two original gradients $g_1, g_2$, are directly clipped to produce clipped gradients $\tilde{g}_1, \tilde{g}_2$ which lie on the surface of a sphere with radius $C$. Noise $n$ is then added into $\tilde{g}_1 + \tilde{g}_2$ and the noisy gradient $\hat{g}$ is obtained. This process is depicted on the left side of Fig. \ref{explaination}. For DP-PSASC, the original gradients $g_1, g_2$ are scaled down to $\tilde{g}_1, \tilde{g}_2$ that also lie on a sphere but with a radius of $\frac{1}{s}$. After summing the gradients $\tilde{g}_1 + \tilde{g}_2$, DP-PSASC first stretches them to $C \cdot (\tilde{g}_1 + \tilde{g}_2)$ and then the noise $n$ is injected into $C \cdot (\tilde{g}_1 + \tilde{g}_2)$ to get the noisy gradient $\hat{g}$. This process is illustrated on the right side of Fig. \ref{explaination}. Given the non-private gradient $g_{\text{np}}$, we can see the angle between $\hat{g}$ in the right part and $g_{\text{np}}$ is likely larger than the angle between $\hat{g}$ in the left part and $g_{\text{np}}$. 

This analysis indicates that direct clipping may provide a weaker privacy guarantee for the directionality of non-private gradients compared to DP-PSASC. Additionally, as the training process nears convergence in its later stages, taking small, slightly inaccurate steps can more effectively optimize model parameters than taking larger, more accurate steps, thereby improving model performance.

\textbf{Gradient scaling weight.} Fig. \ref{scaling_fig} presents the gradient scaling weights for both DP-PSAC and our proposed DP-PSASC methods. It is evident that DP-PSASC assigns larger average weights to smaller gradients during the later stages of training. Additionally, the average weights generally decrease as training progresses across all methods. This trend suggests that gradient norms diminish over time. Furthermore, the decline in average weights helps to prevent anomalous scaled gradients and ensures stable convergence.

\section{Conclusion and future work}
In this paper, we introduce a non-monotonic adaptive scaling weight approach for gradient updates and propose a novel differentially private stochastic gradient descent method, called DP-PSASC. We implement a scaling coefficient $s < 1$ that is easier to search than $C$. Additionally, we present a momentum-enhanced variant of DP-PSASC that addresses the bias introduced by stochastic sampling. We support our methodology with both theoretical analysis and empirical evidence, demonstrating that DP-PSASC surpasses existing DP-SGD methods in performance. Looking forward, we plan to explore the principles for selecting $s$ across various datasets and models. Moreover, there is potential to further enhance the efficiency of DP-PSASC.

\bibliographystyle{plain}
\bibliography{neurips}

\begin{IEEEbiography}[{\includegraphics[width=1in,height=1.25in,clip,keepaspectratio]{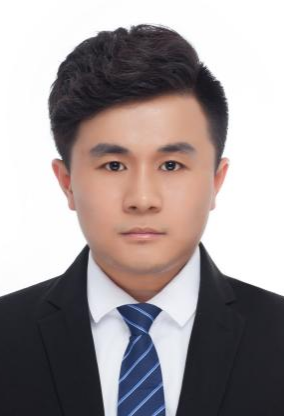}}]{Tao Huang}
	is an Associate Professor at Minjiang University in Fuzhou, Fujian, China. His research interests include deep learning, differential privacy, and computer vision. Huang received his Ph.D in computer science from Renmin University of China. He is a member of IEEE and CCF. Contact him at huang-tao@mju.edu.cn.
\end{IEEEbiography}

\begin{IEEEbiography}[{\includegraphics[width=1in,height=1.25in,clip,keepaspectratio]{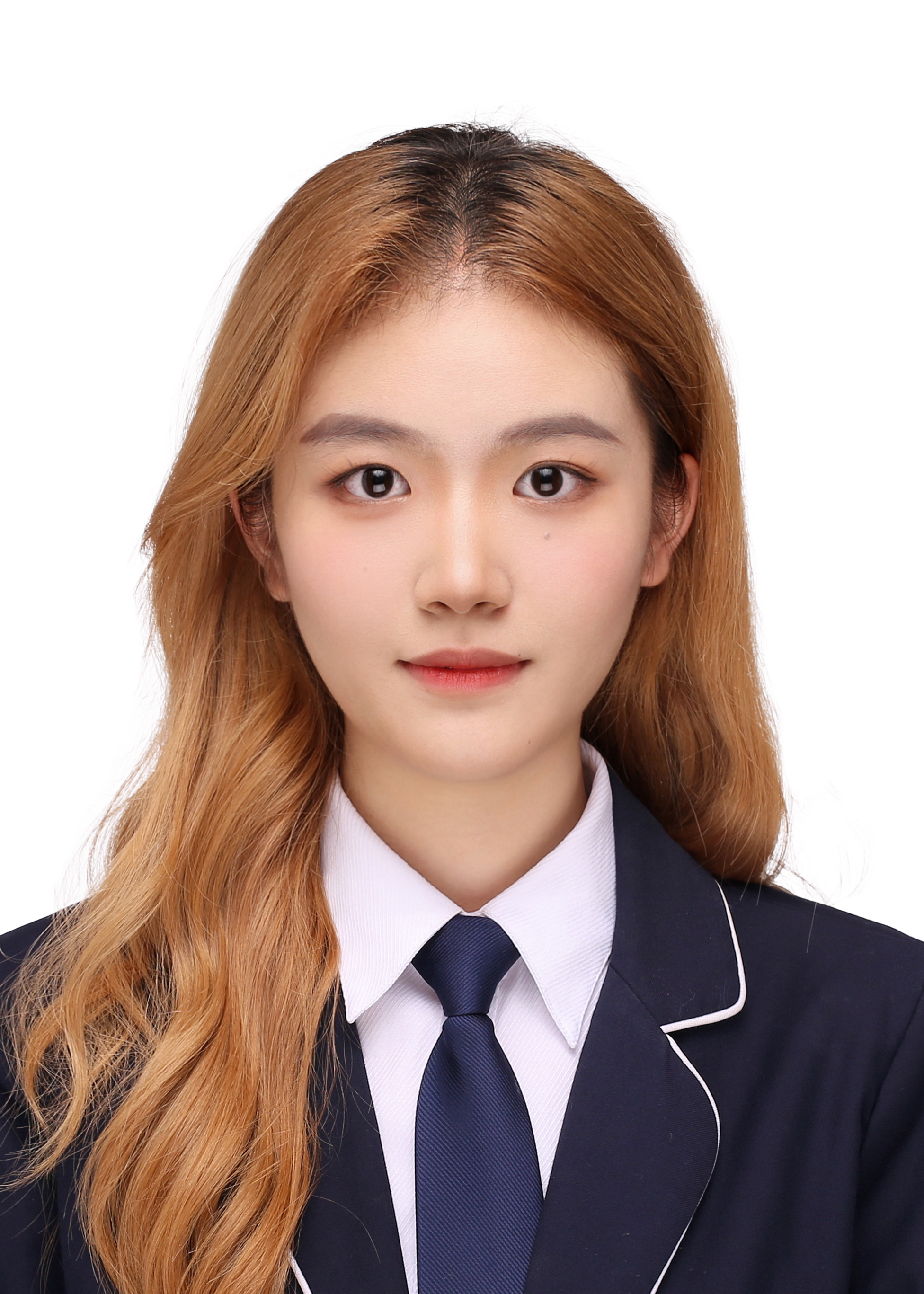}}]{Qingyu Huang}
	is an undergraduate student at Minjiang University in Fuzhou, Fujian, China. Her research interests include differential privacy and machine learning. She is a member of CCF. Contact her at qingyuhuang@stu.mju.edu.cn.
\end{IEEEbiography}

\begin{IEEEbiography}[{\includegraphics[width=1in,height=1.25in,clip,keepaspectratio]{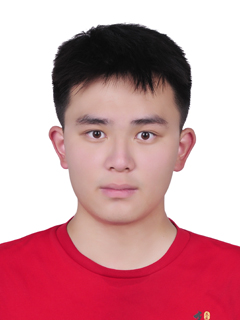}}]{Xin Shi}
	is an undergraduate student at Minjiang University in Fuzhou, Fujian, China. His research interests include differential privacy and machine learning. He is a member of CCF. Contact her at Shixin@stu.mju.edu.cn.
\end{IEEEbiography}

\begin{IEEEbiography}[{\includegraphics[width=1in,height=1.25in,clip,keepaspectratio]{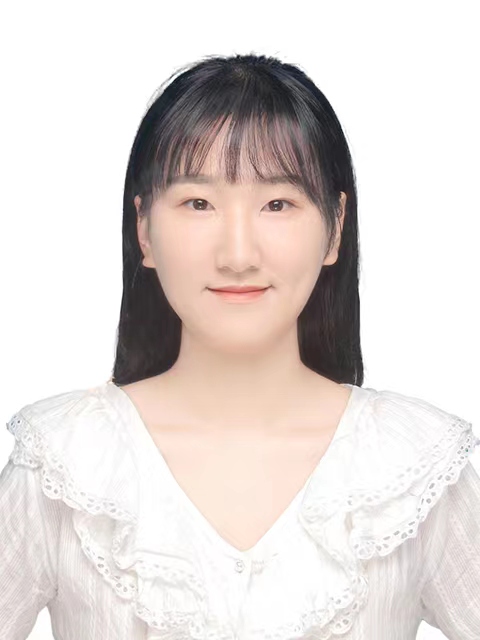}}]{Jiayang Meng}
	is a PhD candidate at School of Information, Renmin University of China. Her research interests include data privacy and security, attacks in machine learning, and differential privacy. Meng received her BSc in software engineering from Huazhong University of Science and Technology. Contact her at jiayangmeng@ruc.edu.cn.
\end{IEEEbiography}

\begin{IEEEbiography}[{\includegraphics[width=1in,height=1.25in,clip,keepaspectratio]{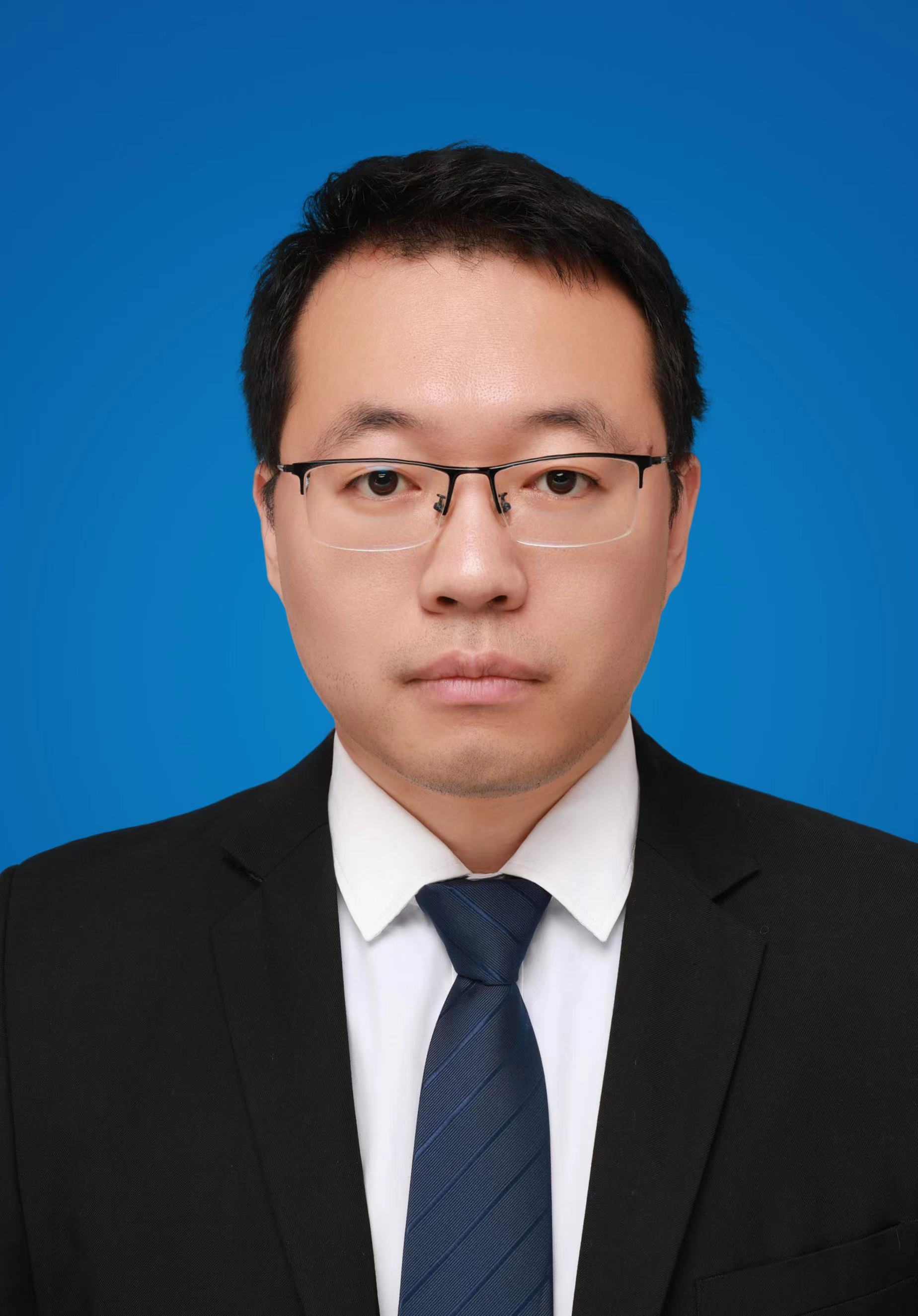}}]{Guolong Zheng}
	is an Associate Professor at Minjiang University in Fuzhou, Fujian, China. His research interests include program analysis, automatic program repair, and AI for software. Zheng received his Ph.D in computer science from University of Nebraska-Lincoln. He is a member of ACM and CCF. Contact him at gzheng@mju.edu.cn.
\end{IEEEbiography}

\begin{IEEEbiography}[{\includegraphics[width=1in,height=1.25in,clip,keepaspectratio]{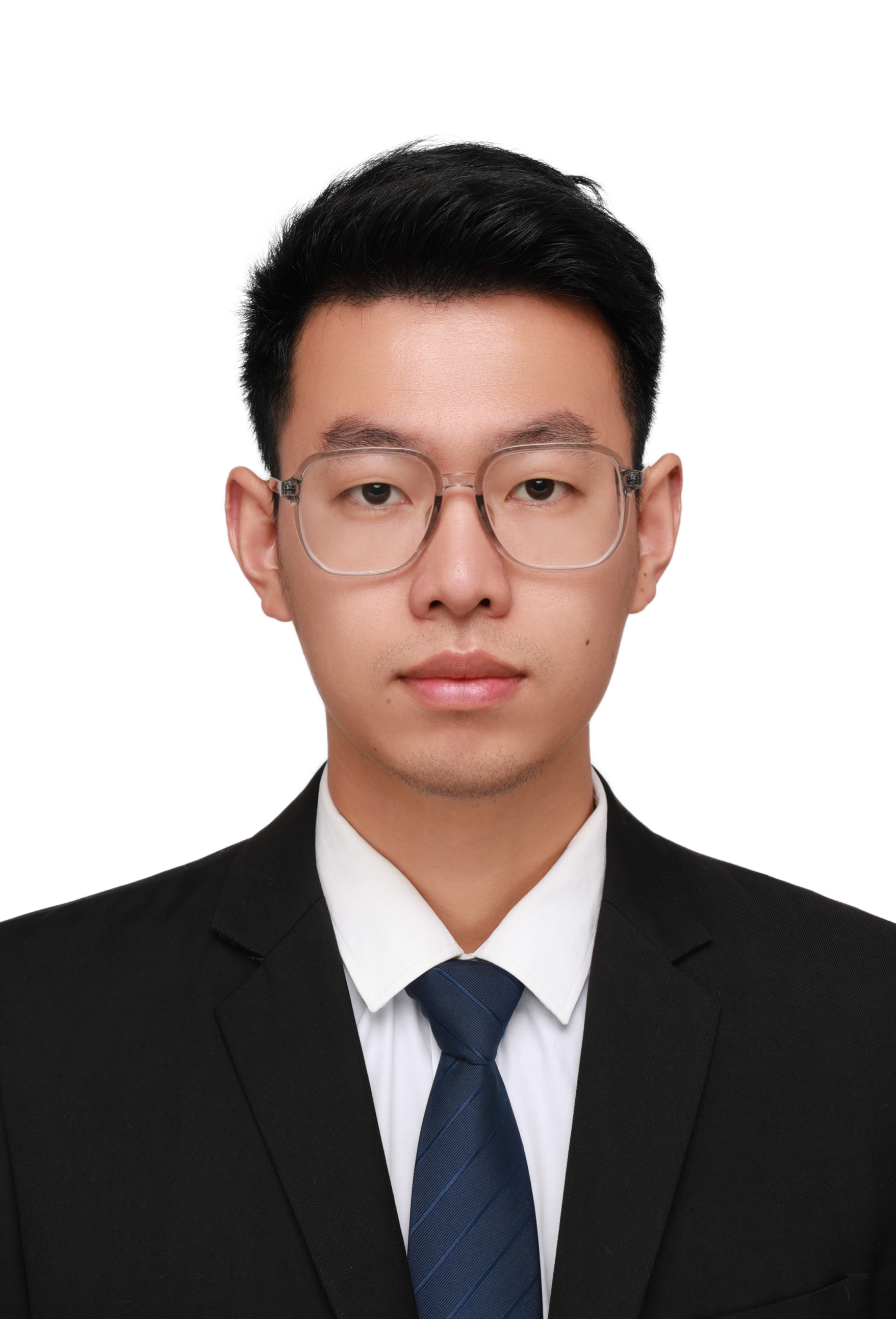}}]{Xu Yang}
	is currently an Associate Professor at the College of Computer and Data Science, Minjiang University, China. He received the Ph.D. degree from the School of Computing Technologies, RMIT University, Australia, with CSIRO's Data61 in 2021, and used to be a Postdoctoral Researcher with the School of Mathematics and Statistics, Fujian Normal University. He has published papers in major conferences/journals, such as IEEE TSC, IEEE TSUSC, IEEE IoTJ, VEH COMMUN, etc. His research interests include cryptography and information security. 
\end{IEEEbiography}

\begin{IEEEbiography}[{\includegraphics[width=1in,height=1.25in,clip,keepaspectratio]{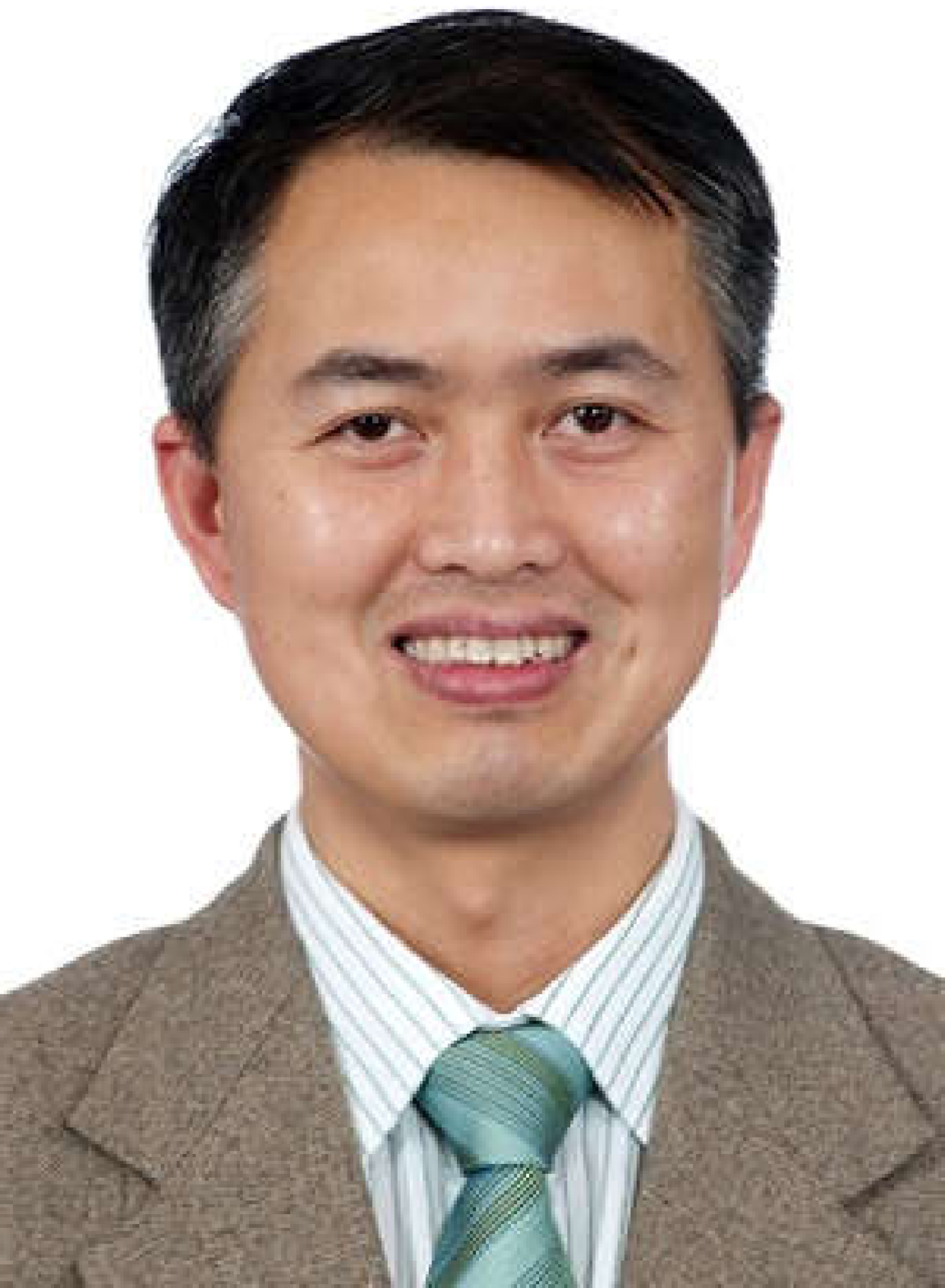}}]{Xun Yi}
	is currently a Professor with the School of Computing Technologies, RMIT University, Australia. His research interests include data privacy protection, Cloud and IoT security, Blockchain, network security and applied cryptography. He has published over 300 research papers. Currently, he is an Associate Editor for IEEE Trans. on Dependable and Secure Computing, IEEE Trans. on Knowledge and Data Engineering, ACM Computing Survey, Information Science (Elsevier) and Journal of Information Security and Application (Elsevier).
\end{IEEEbiography}

\vfill

\end{document}